\pdfoutput=1

\documentclass[11pt]{article}

\usepackage[final]{acl}

\usepackage{times}
\usepackage{latexsym}

\usepackage[T1]{fontenc}

\usepackage[utf8]{inputenc}

\usepackage{microtype}
\usepackage{pifont}
\usepackage{inconsolata}

\usepackage{graphicx}

\usepackage{CJKutf8}
\usepackage{amsmath}
\usepackage{makecell}
\usepackage{array, multirow}
\usepackage[normalem]{ulem}
\useunder{\uline}{\ul}{}
\usepackage{booktabs}
\usepackage{nicematrix}
\usepackage{color, colortbl}
\usepackage{hhline}
\usepackage{rotating}
\usepackage{lscape}
\usepackage{subfigure}
\usepackage{graphicx}
\usepackage{enumitem}
\usepackage{ragged2e}
\usepackage{graphicx}
\usepackage{overpic}
\usepackage[most]{tcolorbox} 
\usepackage{xcolor} 
\usepackage{tipa}

\title{Polishing Every Facet of the GEM\includegraphics[width=1em]{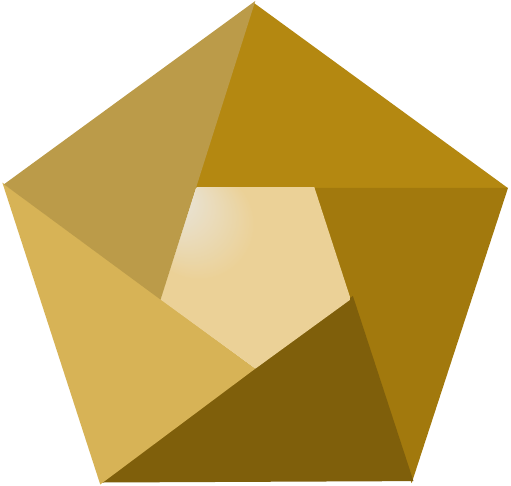}: \\Testing Linguistic Competence of LLMs and Humans in Korean}

\author{
  \textbf{SungHo Kim}\textsuperscript{1}\thanks{\enspace These authors contributed equally to this work.},
  \textbf{Nayeon Kim}\textsuperscript{2}\footnotemark[1],
  \textbf{Taehee Jeon}\textsuperscript{3},
  \textbf{SangKeun Lee}\textsuperscript{1,2}
\\
  \textsuperscript{1}Department of Artificial Intelligence, Korea University, Seoul, South Korea \\
  \textsuperscript{2}Department of Computer Science and Engineering, Korea University, Seoul, South Korea \\
  \textsuperscript{3}Institute for Digital HUSS, Korea University, Seoul, South Korea
\\
  \texttt{\{sungho3268, lilian1208, taeheejeon22, yalphy\}@korea.ac.kr}
}

\begin{document}

\begin{CJK}{UTF8}{mj}

\renewcommand{\ttdefault}{cmtt} 

\maketitle
\begin{abstract}

We introduce the \underline{Ko}rean \underline{G}rammar \underline{E}valuation Bench\underline{M}ark (KoGEM), designed to assess the linguistic competence of LLMs and humans in Korean. KoGEM consists of 1.5k multiple-choice QA pairs covering five main categories and 16 subcategories.
The zero-shot evaluation of 27 LLMs of various sizes and types reveals that while LLMs perform remarkably well on straightforward tasks requiring primarily definitional knowledge, they struggle with tasks that demand the integration of real-world experiential knowledge, such as phonological rules and pronunciation.
Furthermore, our in-depth analysis suggests that incorporating such experiential knowledge could enhance the linguistic competence of LLMs. 
With KoGEM, we not only highlight the limitations of current LLMs in linguistic competence but also uncover hidden facets of LLMs in linguistic competence, paving the way for enhancing comprehensive language understanding.
Our code and dataset are available at:
\href{https://github.com/SungHo3268/KoGEM}{https://github.com/SungHo3268/KoGEM}.
\end{abstract}

\section{Introduction}

Although large language models (LLMs) have demonstrated remarkable performance across various natural language tasks, it is uncertain whether they possess genuine linguistic competence—the ability to understand the underlying principles of a language~\cite{chomsky1965, waldis-etal-2024-holmes}. Their strong performance might stem from their extensive training data rather than understanding of language itself~\cite{bender2021parrot}. Thus, to explore whether LLMs truly understand language beyond statistical pattern recognition, it is crucial to investigate their linguistic competence. However, due to the implicit characteristics of linguistic competence, directly assessing such competence is challenging~\cite{NamKilim2024Generative}.

\begin{figure}[t!]
  \centering
  \includegraphics[width=\linewidth]{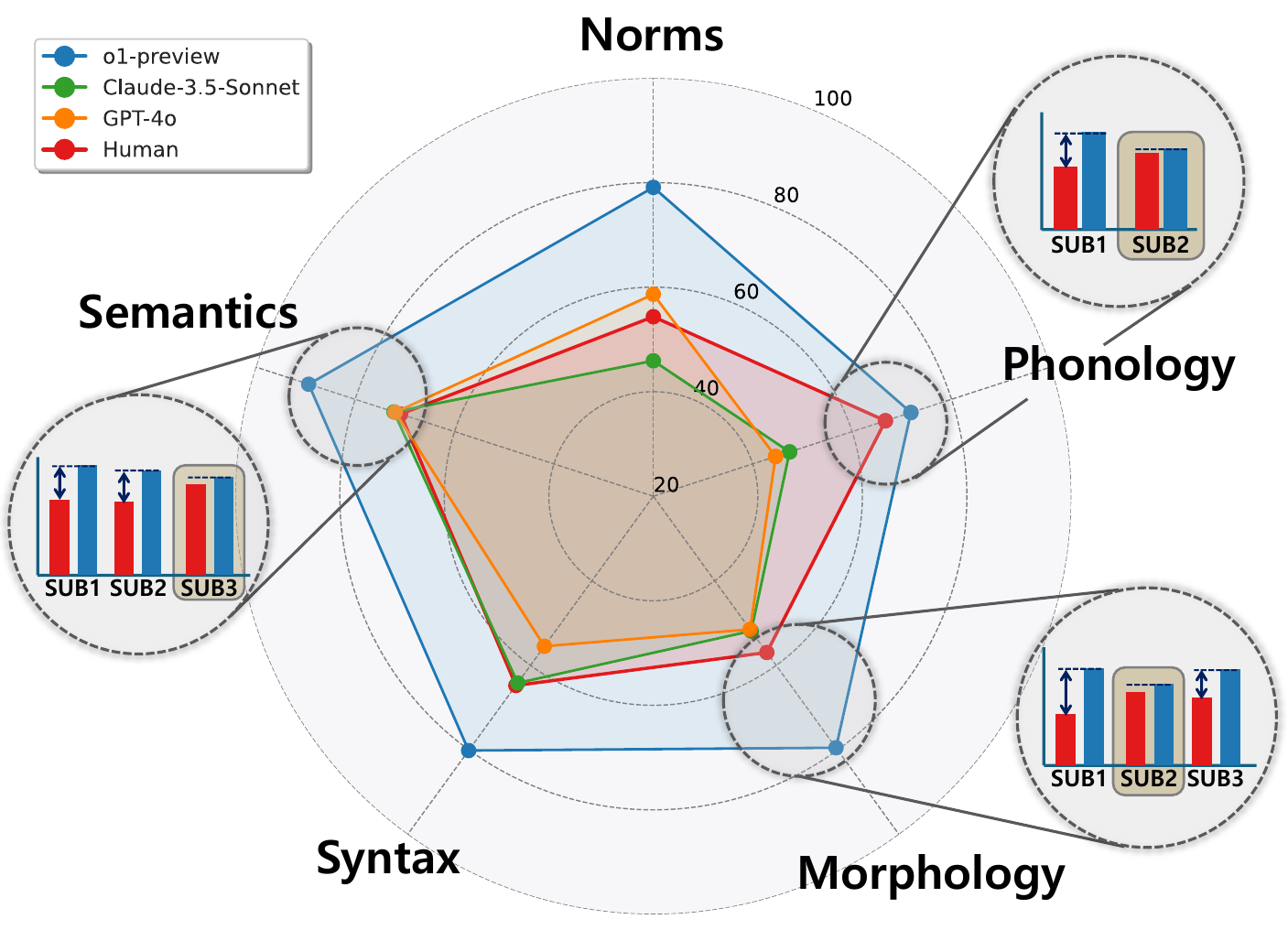}
  \caption{Zero-shot accuracy of the top three LLMs and human performance on KoGEM. Dashed gray circles indicate the accuracy of human (red box) and o1-preview (blue box) in each subcategory. \texttt{SUB} denotes the subcategory corresponding to each main category.}
  \label{figure_1}
\end{figure}

One promising approach to address this challenge is to leverage grammar as a measurable proxy. As grammar explicitly formulates the universal rules of a language~\cite{white1989universal}, it can serve as an effective way to assess the linguistic competence of LLMs. Previous studies have explored the linguistic competence of language models with grammatical knowledge~\cite{hewitt-manning-2019-structural, blevins-etal-2023-prompting, amouyal-etal-2024-large}. 
While various aspects of linguistic knowledge exist, such as phonology and pragmatics, most of the previous works have primarily focused on morphological and syntactic knowledge in English. Additionally, they have paid little attention to other languages, including Korean. 

Since individual languages possess unique linguistic properties, each language should be considered independently to evaluate how well LLMs understand its linguistic knowledge. For this reason, this paper specifically focuses on the Korean language to facilitate a deeper discussion. Unlike English, Korean, as an agglutinative language, exhibits significant morphological variation~\cite{kim-etal-2022-break, kim-etal-2024-kombo, seo-etal-2023-chef}. In addition, the writing system of Korean gives rise to unique phonological rules, such as consonant assimilation and vowel harmony~\cite{cho2020korean}.

With this motivation, we present the \underline{Ko}rean \underline{G}rammar \underline{E}valuation Bench\underline{M}ark, KoGEM, a comprehensive and fine-grained dataset designed to assess the linguistic competence of LLMs in Korean. KoGEM consists of 1,524 multiple-choice grammar questions organized into five main categories—\texttt{Phonology, Morphology, Syntax, Semantics,} and \texttt{Norms}—which are further divided into 16 subcategories based on theoretical linguistics~\cite{Lyons_1968}. This structured taxonomy enables KoGEM to provide a wide and detailed framework for evaluating the linguistic competence of LLMs.

We evaluate humans and a diverse range of open- and closed-source LLMs, including both Korean- and English-centric models, in a zero-shot setting on KoGEM.
Figure~\ref{figure_1} presents the key results, comparing human performance with the top three LLMs on KoGEM.
At first glance, the outstanding performance of o1-preview may appear flawless, suggesting that it surpasses humans in all aspects. However, a closer analysis of linguistic phenomena, breaking down major linguistic categories into finer subcategories, reveals highly varied tendencies. Notably, we identify certain hidden facets where humans perform relatively well while LLMs lag behind, highlighting potential areas for improvement.
Through an in-depth analysis of these hidden facets, we demonstrate substantial improvements when LLMs are augmented with the experiential knowledge that humans naturally acquire through real-world experience.
In summary, our contributions are as follows:
\begin{itemize}[left=0.3cm]
    \item We introduce KoGEM, a comprehensive and fine-grained benchmark designed to objectively assess Korean grammatical knowledge based on theoretical linguistics.
    \item We evaluate 27 open- and closed-source LLMs, including Korean- and English-centric models, across 16 fine-grained Korean grammar taxonomy, comparing them with humans. 
    \item We reveal novel insights into the strengths and limitations of LLMs through in-depth analysis, paving the way for enhancing LLMs and addressing gaps in their linguistic competence.    
\end{itemize}

\begin{figure}[t!]
    \includegraphics[width=1.0\linewidth]{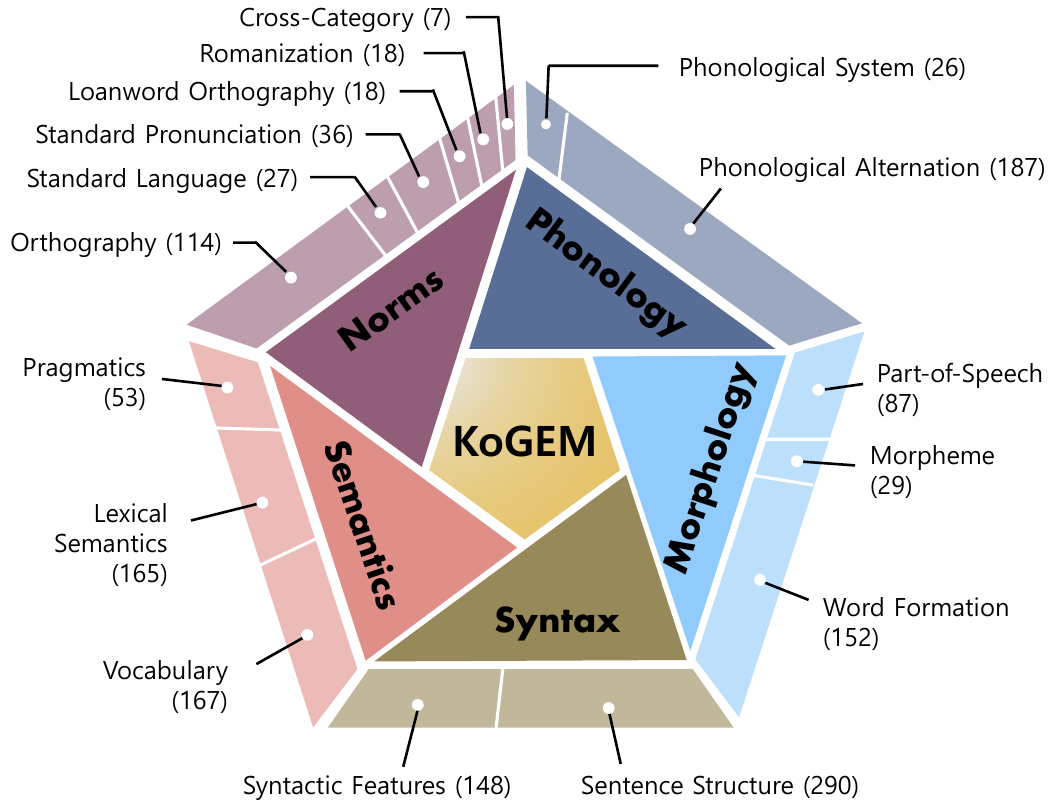}
    \caption{Data distribution of KoGEM, categorized into five main categories and 16 subcategories in total. The number in parentheses next to each subcategory name represents the number of questions it contains.}
    \label{fig3:data_distribution}
    \vspace{-5pt}
\end{figure}

\section{Korean Grammar Taxonomy}
\label{sec:KGT}

Before delving into the details, we define the concept of `grammar.' This paper focuses solely on prescriptive grammar due to the extensiveness of descriptive grammar. While prescriptive grammar does not encompass all aspects of a language's grammatical system, it is firmly based on foundational knowledge of grammar. By examining the degree to which prescriptive grammar is understood, we can evaluate the linguistic competence of both LLMs and humans.

To assess linguistic competence, we set the categories of grammar as the main fields of theoretical linguistics: phonology, morphology, syntax, and semantics. 
While pragmatics can be treated as a separate field, we include it under semantics in this framework due to its close relationship with semantics and the relative absence of pragmatics considerations in prescriptive grammar. Additionally, we include linguistic norms as a category, reflecting one of the core principles of prescriptive grammar. Although norms cannot inherently capture the diversity of linguistic phenomena, understanding norms requires linguistic competence, which is the focus of our evaluation.

Further, we define the subcategories for each main linguistic category. The subcategories encompass key subfields of each main category and are aligned with the structure of the current Korean high school education system. For example, in phonology, the subfields regarding the phonological system and variation in Korean are divided into phonological system and phonological alternation. Further details can be found in Appendix \ref{sec:appendix_taxonomy}. 

\begin{itemize}[left=0.3cm]
    \item \textbf{Phonology}: Phonological System, Phonological Alternation 
    \item \textbf{Morphology}: Part-of-Speech, Morpheme, Word Formation 
   \item \textbf{Syntax}: Sentence Structure, Syntactic Features 
   \item \textbf{Semantics}: Vocabulary, Lexical Semantics, Pragmatics
   \item \textbf{Norms}: Orthography, Standard Language, Standard Pronunciation, Loanword Orthography, Romanization, Cross-Category
\end{itemize}

\section{KoGEM}
The primary purpose of our dataset is to assess the linguistic competence of LLMs and humans. To this end, we introduce KoGEM, a \textbf{Ko}rean \textbf{G}rammar \textbf{E}valuation bench\textbf{M}ark, containing 1.5k grammar question-answer (QA) pairs, categorized into five main categories and 16 subcategories based on a predefined taxonomy described in Section~\ref{sec:KGT}.

\subsection{Dataset Construction}
\label{sec:dataset_construction}
We provide a detailed explanation of source data, data format, collection, and categorization. 

\paragraph{1) Source Data}
To encompass the Korean grammar taxonomy defined in Section~\ref{sec:KGT}, we extract Korean grammar questions from four types of official exams: (1) the College Scholastic Ability Test (CSAT); (2) the National United Achievement Test (NUAT); (3) the High School Qualification Exam (HSQE); and (4) the Civil Service Exam (CSE). 
While other Korean language tests exist, we especially selected these exams as they are designed for native Korean speakers, ensuring that the questions reflect linguistic competence expected in academic and professional contexts.
A specific description of our source data is provided in Appendix~\ref{sec:appendix_source_data}.

\paragraph{2) Data Format}
As prescriptive grammar emphasizes the correct use of language, we design our task as a multiple-choice QA task with a clearly defined correct answer. Each QA pair consists of up to four components: {\texttt{passage, question, paragraph,} and \texttt{choices}}. The \texttt{passage} provides the necessary context to understand the question, while the \texttt{paragraph} offers brief explanations or examples of grammatical concepts relevant to answering it. The \texttt{choices} include either four or five answer options, depending on the type of exam. Examples of QA pairs can be found in Figures~\ref{fig:example_phonology} to \ref{fig:example_norms}.

\paragraph{3) Data Collection}
As the source data is publicly available in PDF format, we extract the text using optical character recognition (OCR). After OCR, three authors manually review every grammar question and apply preprocessing steps such as underlining, segmentation, and box separation using Hypertext Markup Language (HTML) formatting. Questions primarily relying on images for context are excluded, as we evaluate LLMs solely in a text-based, single-modal setting. Additionally, tables in the data are converted into sequential text based on a set of predefined rules. Detailed dataset preprocessing is described in Appendix~\ref{sec:appendix_preprocessing}.

\paragraph{4) Data Categorization}
To assess the linguistic competence of LLMs in various grammatical areas, three Korean language majors categorize the preprocessed questions into 16 subcategories. Each annotator independently classifies each question, first identifying the main linguistic category and then assigning the appropriate subcategory. To ensure the reliability of our categorization, labels are finalized by majority vote, and disagreements are resolved through discussion. Questions overlapping two linguistic categories are classified into one category based on the context of the correct answer. Additionally, we remove questions requiring knowledge from more than three categories to focus on the evaluation of each linguistic subcategory.

\subsection{Data Statistics}
\label{sec:data_statistics}
The finalized KoGEM benchmark consists of a total of 1,524 annotated QA pairs. Figure~\ref{fig3:data_distribution} illustrates the distribution of KoGEM across each main category and subcategory, classified according to the Korean grammar taxonomy. Moreover, we present the specific number of samples for each main category from each source exam in Figure~\ref{fig:subcategory_numbers}.

\begin{table*}[t!]
    \setlength{\tabcolsep}{5pt}             
    \renewcommand{\arraystretch}{1.00}       
    \small
    \begin{tabular}{ccl|cccccc}
        \toprule
        \multicolumn{1}{c|}{Language} & \multicolumn{1}{c|}{Type} & \multicolumn{1}{c|}{Model} & Phonology & Morphology & Syntax & Semantics & Norm & Avg. \\ 
        \midrule
        \multicolumn{1}{c|}{\multirow{8}{*}{Korean}} & \multicolumn{1}{c|}{\multirow{6}{*}{Open}} & Bllossom-8B & 24.41 & 25.00 & 22.15 & 34.55 & 29.09 & 27.10 \\
        \multicolumn{1}{c|}{} & \multicolumn{1}{c|}{} & SOLAR-v1.0-10.7B-Instruct & 24.88 & 25.75 & 26.71 & 31.69 & 25.45 & 27.36 \\
        \multicolumn{1}{c|}{} & \multicolumn{1}{c|}{} & KULLM-3-10.7B & 21.60 & 26.49 & 26.03 & 29.35 & 28.18 & 26.64 \\
        \multicolumn{1}{c|}{} & \multicolumn{1}{c|}{} & EEVE-v1.0-10.8B-Instruct & 22.54 & 27.24 & 27.85 & 40.78 & 27.73 & 30.25 \\
        \multicolumn{1}{c|}{} & \multicolumn{1}{c|}{} & EXAONE-3.5-7.8B-Instruct & 24.88 & 30.22 & 32.19 & 43.64 & 31.36 & 33.60 \\
        \multicolumn{1}{c|}{} & \multicolumn{1}{c|}{} & EXAONE-3.5-32B-Instruct & 27.23 & 37.31 & 36.30 & 50.65 & 37.27 & 38.98 \\
        \cmidrule{2-9}
        \multicolumn{1}{c|}{} & \multicolumn{1}{c|}{\multirow{2}{*}{Closed}} & HyperCLOVA-HCX-DASH-001 & 23.94 & 31.34 & 25.57 & 39.74 & 31.36 & 30.77 \\
        \multicolumn{1}{c|}{} & \multicolumn{1}{c|}{} & HyperCLOVA-HCX-003 & 32.39 & 41.79 & 41.10 & 55.32 & 48.18 & 44.62 \\ 
        \midrule
        \multicolumn{1}{c|}{\multirow{19}{*}{English}} & \multicolumn{1}{c|}{\multirow{10}{*}{Open}} & Gemma-2-9B-Instruct & 24.41 & 31.34 & 31.05 & 42.08 & 29.09 & 32.68 \\
        \multicolumn{1}{c|}{} & \multicolumn{1}{c|}{} & Gemma-2-27B-Instruct & 27.33 & 29.48 & 36.76 & 47.27 & 29.09 & 35.70 \\
        \multicolumn{1}{c|}{} & \multicolumn{1}{c|}{} & Qwen2.5-7B-Instruct & 23.00 & 30.97 & 34.02 & 44.42 & 25.00 & 33.27 \\
        \multicolumn{1}{c|}{} & \multicolumn{1}{c|}{} & Qwen2.5-14B-Instruct & 29.58 & 38.43 & 39.27 & 52.21 & 33.64 & 40.22 \\
        \multicolumn{1}{c|}{} & \multicolumn{1}{c|}{} & Qwen2.5-32B-Instruct & 26.29 & 36.19 & 41.78 & 62.86 & 35.45 & 43.04 \\
        \multicolumn{1}{c|}{} & \multicolumn{1}{c|}{} & DeepSeek-R1-Distill-Qwen-14B & 29.11 & 32.09 & 39.27 & 47.79 & 32.27 & 37.73 \\
        \multicolumn{1}{c|}{} & \multicolumn{1}{c|}{} & DeepSeek-R1-Distill-Qwen-32B & 36.15 & 45.52 & 40.41 & 61.56 & 30.00 & 44.55 \\
        \multicolumn{1}{c|}{} & \multicolumn{1}{c|}{} & s1-32B & 39.91 & 43.28 & 46.12 & 62.60 & 40.00 & 48.03 \\
        \multicolumn{1}{c|}{} & \multicolumn{1}{c|}{} & Llama-3.1-8B-Instruct & 24.41 & 24.25 & 23.97 & 36.36 & 26.82 & 27.62 \\
        \multicolumn{1}{c|}{} & \multicolumn{1}{c|}{} & Llama-3-70B & 24.41 & 32.46 & 34.47 & 45.71 & 27.73 & 34.58 \\
        \multicolumn{1}{c|}{} & \multicolumn{1}{c|}{} & Llama-3.1-405B & 37.56 & 43.28 & 47.03 & 62.34 & 35.00 & 47.18 \\
        \cmidrule{2-9}
        \multicolumn{1}{c|}{} & \multicolumn{1}{c|}{\multirow{8}{*}{Closed}} & Gemini-1.5-flash & 37.56 & 38.81 & 44.52 & 60.78 & 35.45 & 45.34 \\
        \multicolumn{1}{c|}{} & \multicolumn{1}{c|}{} & Gemini-2.0-flash-exp & 46.01 & 49.63 & 56.85 & 70.91 & 45.91 & 56.04 \\
        \multicolumn{1}{c|}{} & \multicolumn{1}{c|}{} & Claude-3-haiku & 21.13 & 34.33 & 35.62 & 44.68 & 30.91 & 34.97 \\
        \multicolumn{1}{c|}{} & \multicolumn{1}{c|}{} & Claude-3.5-Sonnet & {\ul 47.42} & {\ul 52.61} & {\ul 64.38} & {\ul 74.55} & 46.82 & {\ul 59.97} \\
        \multicolumn{1}{c|}{} & \multicolumn{1}{c|}{} & GPT-3.5-turbo & 21.13 & 27.61 & 27.40 & 31.43 & 25.45 & 27.30 \\
        \multicolumn{1}{c|}{} & \multicolumn{1}{c|}{} & GPT-4o-mini & 32.39 & 34.70 & 36.53 & 52.21 & 39.09 & 39.96 \\
        \multicolumn{1}{c|}{} & \multicolumn{1}{c|}{} & GPT-4o & 44.60 & 51.49 & 55.48 & 71.95 & {\ul 58.64} & 57.87 \\
        \multicolumn{1}{c|}{} & \multicolumn{1}{c|}{} & o1-preview & \textbf{71.83} & \textbf{79.48} & \textbf{80.14} & \textbf{89.35} & \textbf{79.09} & \textbf{81.04} \\ 
        \midrule
        \multicolumn{3}{c|}{LLMs Avg.} & 31.33 & 37.08 & 39.00 & 51.36 & 35.71 & 40.24 \\ 
        \midrule
        \multicolumn{3}{c|}{Human} & 66.70 & 56.95 & 64.75 & 70.84 & 54.34 & 63.04 \\ 
        \bottomrule
    \end{tabular}
    \caption{Zero-shot accuracy evaluation results on our KoGEM benchmark. It consists of four segments: Korean-centric LLMs trained mainly on Korean data, English-centric LLMs trained primarily on English data, and the average accuracy performance of all LLMs and humans, respectively.}
    \label{table_per_category}
\end{table*}

\section{Experiments}
In this section, we evaluate the linguistic competence of both LLMs and humans using KoGEM. Section~\ref{llm_evaluation} describes the experimental settings for the LLMs in the view of baselines and evaluation metrics, while Section~\ref{human_evaluation} provides a detailed explanation of the methods for assessing human performance. Finally, Section~\ref{main_results} presents comprehensive experimental results for each main category.

\subsection{LLM Evaluation}
\label{llm_evaluation}
\paragraph{Baselines.}
\label{baselines}
We evaluated 8 Korean-centric LLMs, such as the EXAONE and HyperCLOVA X series, as well as 19 well-known English-centric LLMs, including the OpenAI-GPT, Claude, and Gemini series. Specifically, we compared open-source LLMs of various sizes, ranging from smaller models with 8B parameters to cutting-edge closed-source models such as o1-preview. Detailed descriptions of the evaluated models can be found in Appendix~\ref{llm_details}.

\paragraph{Evaluation Metrics.}
\label{evaluation_metrics}
To assess the Korean grammatical knowledge of LLMs on KoGEM, we measure accuracy through zero-shot evaluation and few-shot evaluation.\footnote{While, in the main body of this paper, we just handle the zero-shot, you can find the few-shot evaluation results in Appendix~\ref{appx:few-shot_results}.} The prompt was structured in the same order as the test presented to humans: \texttt{passage}, \texttt{question}, \texttt{paragraph}, and \texttt{choices}, and provided as a single prompt. 
The LLMs were instructed to select the correct answer from the choices and generate a short explanation for their choices. Detailed prompt designs and experimental settings, such as hyperparameters and devices, are provided in Appendix~\ref{appendix: details_of_llm_evaluation}.

\subsection{Human Evaluation}
\label{human_evaluation}
Since the data comes from official competency exams in Korea, we first explored publicly available statistics on human performance. Through this investigation, we obtained average response rates for each question from 10,000+ responses for CSAT and NUAT exams.\footnote{The Korean private education company Megastudy has published accuracy for each question since 2016 (https://www.megastudy.net/Entinfo/correctRate/main.asp).} However, for the other two types of exams, HSQE and CSE, there are no publicly available data for human performance. To gain the human scores for these exams, we conducted crowdsourcing via a data research platform~\footnote{Macromill Embrain, a Korean company specializing in online research (https://embrain.com).}.
To ensure suitable participants for each exam, we recruited different groups of crowdworkers based on the specific characteristics of each exam. Additionally, we gathered 10+ responses per each question and used the average scores. Details of crowdsourcing are described in Appendix~\ref{appendix: human_evaluation}.

\subsection{Results for Main Category}
\label{main_results}
Table~\ref{table_per_category} presents the main linguistic category-specific performances of LLMs and humans on KoGEM. Overall, LLM results generally follow the scaling law~\cite{kaplan-2020-scaling}. Notably, among LLMs, o1-preview was the only model to outperform humans, exceeding their performance by an average of 18.00\%. All other models fell short: Claude-3.5-Sonnet and GPT-4o scored 3.07\% and 5.17\% lower than humans, respectively.

We categorized LLMs into Korean-centric and English-centric groups based on their primary training language. Many Korean-centric models exhibited lower performance, likely due to their smaller sizes and outdated architectures. Surprisingly, the English-centric models s1-32B, DeepSeek-R1-Distill-Qwen-32B, and Qwen2.5-32B, despite their relatively small sizes, outperformed or matched all Korean-centric models. Their strong performance likely stems from their multilingual training across over 29 languages~\cite{qwen2.5}, leveraging shared linguistic features~\cite{chen-etal-2019-multi-source, wang-etal-2024-probing-emergence, chang-etal-2022-geometry}. Moreover, results from o1-preview, s1-32B, and two DeepSeek-R1 series suggest that test-time scaling~\cite{snell2024scalingllmtesttimecompute, s1-32b-stanford-2025} effectively enhances linguistic competence in Korean.

A notable finding is the variation in LLMs and human performance across linguistic categories. Specifically, humans excelled in the Phonology category, which requires multimodal reasoning, such as integrating text with implicit phonological rules and pronunciation. In contrast, the Phonology was the weakest category for LLMs.
Even o1-preview, despite its overall superiority, surpassed humans by only 5.13\% in this category, much lower than the 18.00\% gap observed across all categories.
Conversely, the smallest performance gap between humans and LLMs was observed in the Norms category, which relies heavily on rote knowledge, such as correct spelling and loanword usage.

However, we question whether trends at the main category level are sufficient to fully capture the core linguistic differences between LLMs and humans. To gain deeper insights, Sections~\ref{analysis_subcategory} and \ref{indepth_analysis_subcategory} further break down the five main categories into 16 subcategories. This fine-grained analysis provides a more detailed understanding of LLMs’ linguistic competence, highlighting both their strengths and limitations.

\section{Results for Each Subcategory}
\label{analysis_subcategory}
\begin{figure*}[t!]
    \includegraphics[width=1.0\linewidth]{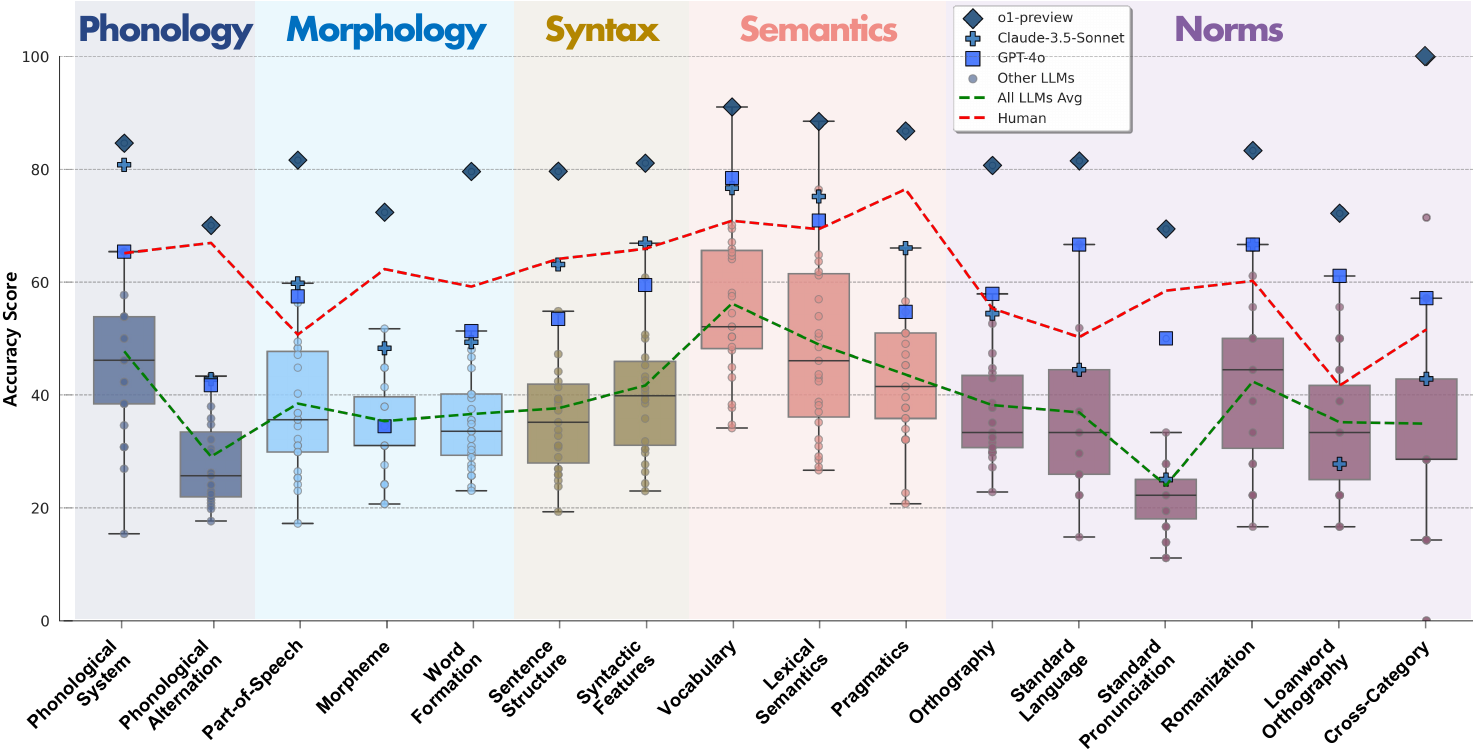}
    \caption{Comparison of LLMs and humans across 16 subcategories. The distributions of all LLMs are depicted using box plots, while individual scores for the top three models are highlighted as scatter plots. A green dashed line indicates the average performance of all LLMs.~\protect\footnotemark, and a red dashed line represents human performance.}
    \label{fig:subcategory_results}
\end{figure*}

Figure~\ref{fig:subcategory_results} compares the performance of LLMs and humans across 16 subcategories. A closer examination of individual subcategories reveals distinct strengths and weaknesses, as LLMs and humans excel in different areas. This underscores the need for a fine-grained evaluation of linguistic competence at the subcategory level.

\paragraph{Phonology}
\label{phonology_analysis}
The phonology category represents the area with the most significant performance gap between LLMs and humans. 
Specifically, regarding the \texttt{Phonological Alternation} subcategory, humans outperformed the average performance of LLMs by over 35\%, marking the largest performance gap among all subcategories of KoGEM. 
We suspect that this performance gap between humans and LLMs stems from the ability of humans to ground multimodal knowledge~\cite{smith1961psychology, Holler-2019}.
Humans intuitively recall the pronunciation of a word~\cite{psychology_carroll1986}, recognizing implicit phonological changes and integrating the pronunciation with the text.
In contrast, LLMs rely on textual input, lacking exposure to spoken language and phonological processing in real-world contexts. This fundamental challenge can limit the ability of LLMs to process phonological alternations effectively.

\paragraph{Morphology} 
Interestingly, humans tend to underperform in the \texttt{Part-of-Speech} subcategory, whereas LLMs achieve their highest performance within the subcategories of the Morphology category. This discrepancy may stem from differences in how humans and LLMs process language. Humans rely on intuitive contextual understanding~\cite{mac-2020-humanlevel} rather than explicitly distinguishing part-of-speech. In contrast, LLMs classify them through data-driven pattern detection, giving them a distinct advantage.
On the other hand, in the \texttt{Morpheme} subcategory, the gap between humans and LLMs is the largest within the Morphology category. 
Humans instinctively decompose words into inflectional morphemes and root words that carry the core meaning, allowing for intuitive interpretation~\cite{marslen2007morphology}. However, LLMs process words either as whole units or as tokenized segments based on their defined vocabulary, making them less suited to handling diverse morphological variations. 

\footnotetext{More detailed top-k average results of LLMs, along with their distributions by subcategory, are provided in Appendix~\ref{appdx:cum_zero_shot_results}.}

\paragraph{Syntax}
LLMs and humans exhibited similar tendencies, particularly showing better performance on \texttt{Syntactic Feature} subcategory than on \texttt{Sentence Structure} subcategory. This trend may be attributed to the unique syntactic characteristics in Korean, which has relatively flexible word order compared to English~\cite{cho2020korean}. For example, the Korean translation of ``I eat rice'' is ``나는 밥을 먹는다\textsubscript{*I rice eat}''. However, this sentence can also appear as ``나는 먹는다 밥을\textsubscript{I eat rice}'' in Korean. This flexibility is difficult to codify, requiring intuitive understanding through real-world experience to comprehend it. 
On the other hand, since \texttt{Syntactic Feature} task follows well-defined rules, such as tense, voice, and honorifics, this tends to be more standardized than the \texttt{Sentence Structure} task.

\begin{figure*}[t!]
    \centering
    \subfigure[Phonology (\texttt{Phonological Alternation})]{\includegraphics[width=0.49\textwidth]{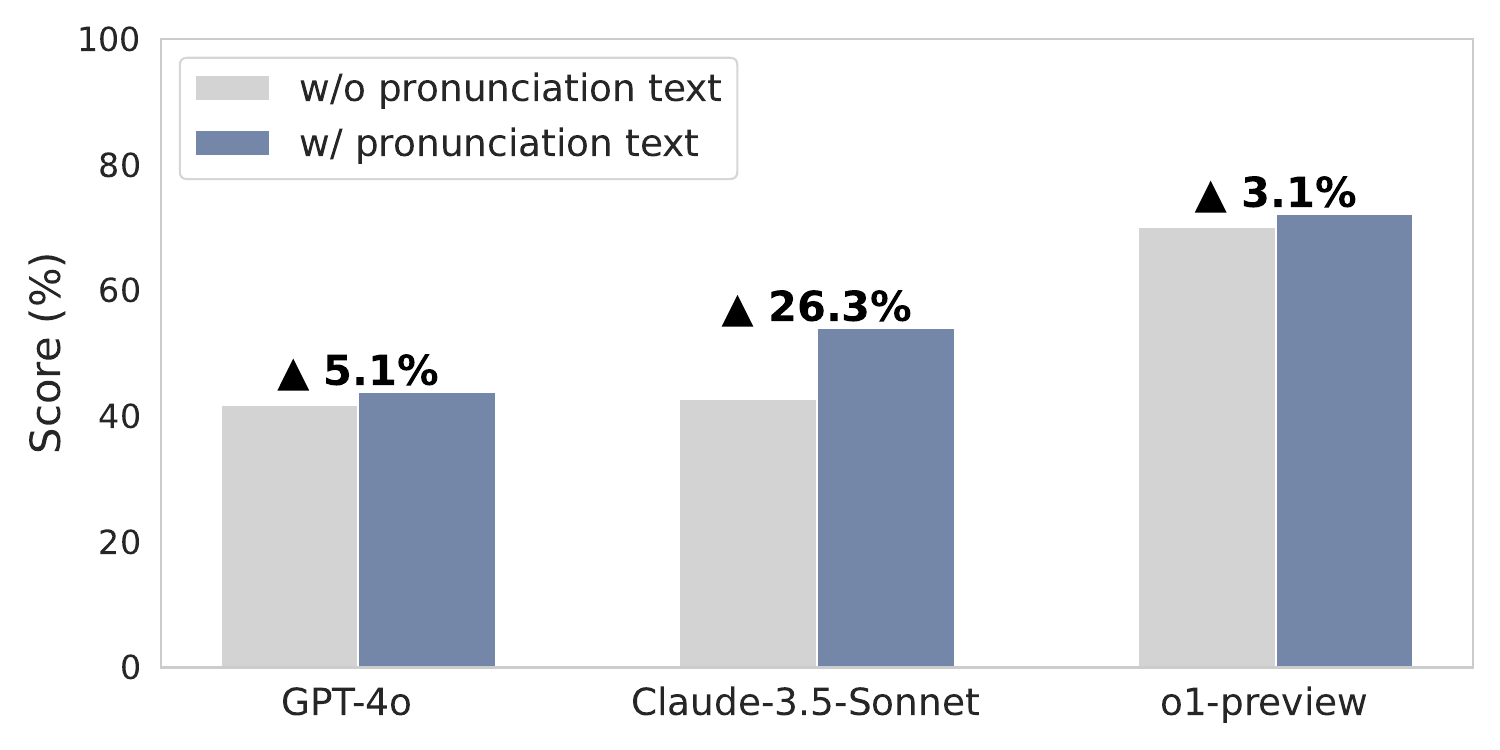}}
    \subfigure[Morphology (\texttt{Morpheme})]{\includegraphics[width=0.49\textwidth]{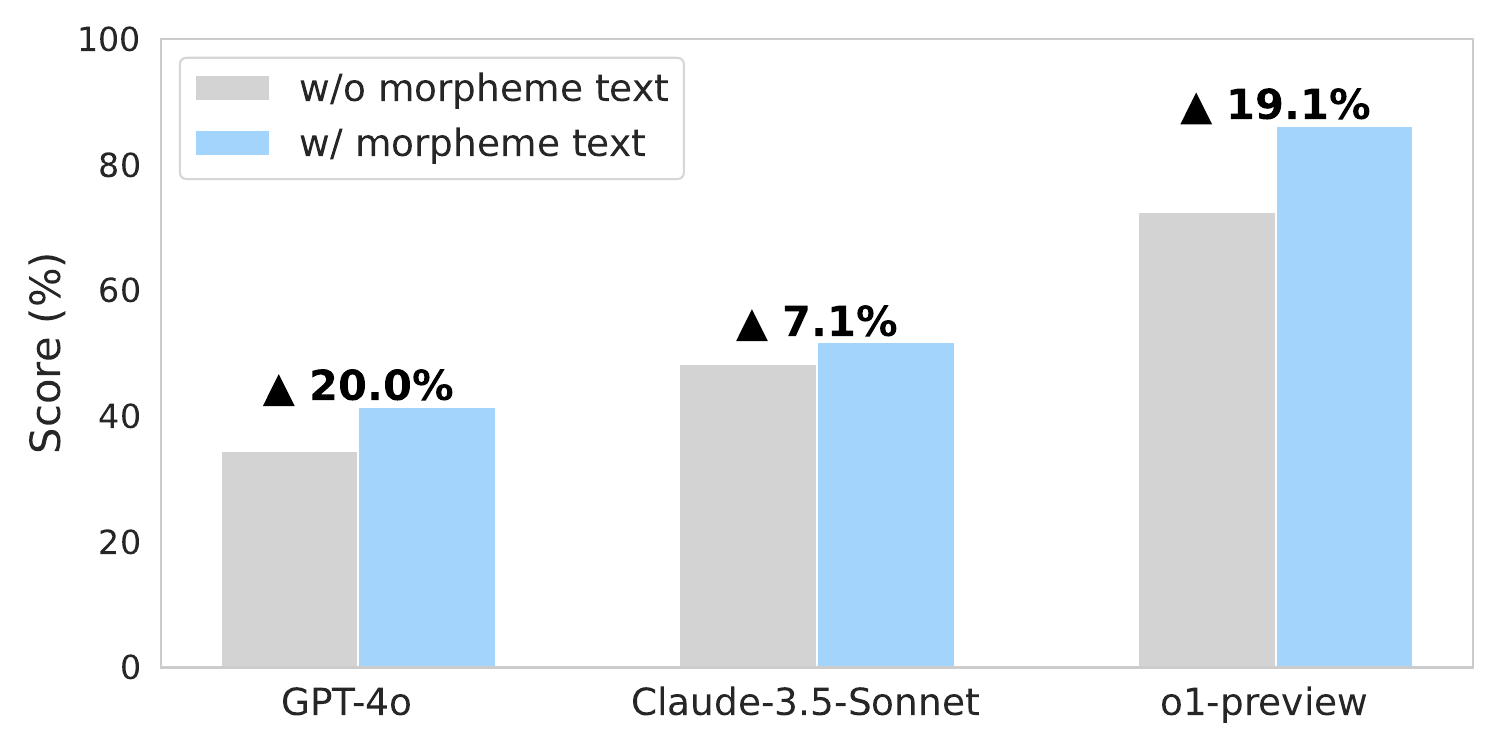}}
    \caption{Comparison of performances in (a) \texttt{Phonological Alternation} and (b) \texttt{Morpheme} subcategories, with and without the additional knowledge, such as \textit{pronunciation text} and \textit{morpheme text}. The values between each bar graph pair indicate the rate of increase(\scalebox{0.8}{\ding{115}}) compared to the original results.}
    \label{fig:exp_knowledge}
    \vspace{-0.1cm}
\end{figure*}

\paragraph{Semantics}
Within the Semantics category, the differences among subcategories are particularly intriguing. Specifically, \texttt{Pragmatics} subcategory shows the largest performance gap within the Semantics category, with an average difference of 32.85\% between LLMs and humans, whereas \texttt{Vocabulary} subcategory exhibits the smallest disparity, averaging 14.66\%.  
This contrast likely stems from the nature of each subcategory. First, in \texttt{Vocabulary} subcategory, many QA tasks involve definitional knowledge and the correct use of words, often requiring straightforward memorization of word meanings. In contrast, \texttt{Pragmatics} subcategory demands an understanding of context-specific intent, such as the relationship between speaker and listener, and the communicative purpose of an utterance, which relies heavily on real-world conversational experience~\cite{Levinson_1983}. As a result, LLMs can struggle with \texttt{Pragmatics} subcategory more than with the relatively straightforward \texttt{Vocabulary} subcategory.

\paragraph{Norms}
Both humans and LLMs exhibited lower overall performance in the Norms category compared to other areas. Although Koreans receive standardized education on linguistic norms as part of their high school curriculum, they often struggle to adhere to standard language regulations~\cite{NIKL_2020}. We suspect that LLMs may have limited exposure to texts explicitly describing standard regulations. As a result, both humans and LLMs may generally exhibit lower performance in the Norms category. Among the subcategories within Norms category, the largest performance gap between LLMs and humans, approximately 34.49\%, was observed in \texttt{Standard Pronunciation}. This trend parallels findings in the Phonology category, suggesting that the gap may stem not only from normative factors but also from phonological features.

\paragraph{Summary}
LLMs exhibit a relatively smaller performance gap compared to humans in tasks that rely on memorization and pattern recognition. In contrast, they fall significantly behind in tasks requiring experiential knowledge, such as \texttt{Phonological Alternation}, \texttt{Morpheme}, and \texttt{Pragmatics}, revealing a substantial performance gap between LLMs and humans. This underscores a key limitation in LLMs' ability to ground linguistic knowledge and real-world experience.

\section{In-depth Analysis for Subcategory}
\label{indepth_analysis_subcategory}
From our previous analysis of subcategory results, we identified certain subcategories that are relatively easy for humans but challenging for LLMs. In this section, we further investigate these specific subcategories. Section~\ref{comparison_time} compares the \textit{thinking} time required to solve problems across all subcategories using the s1-32B, which demonstrated the best performance among open-sourced models employing test-time scaling. Section~\ref{additional_knowledge} examines the impact of incorporating experiential knowledge typically utilized by humans, such as pronunciation and morphemes in \texttt{Phonological Alternation} and \texttt{Morpheme} subcategories, respectively.

\subsection{Comparison of Thinking Time}
\label{comparison_time}
\begin{figure}[t!]
    \includegraphics[width=1.0\linewidth]{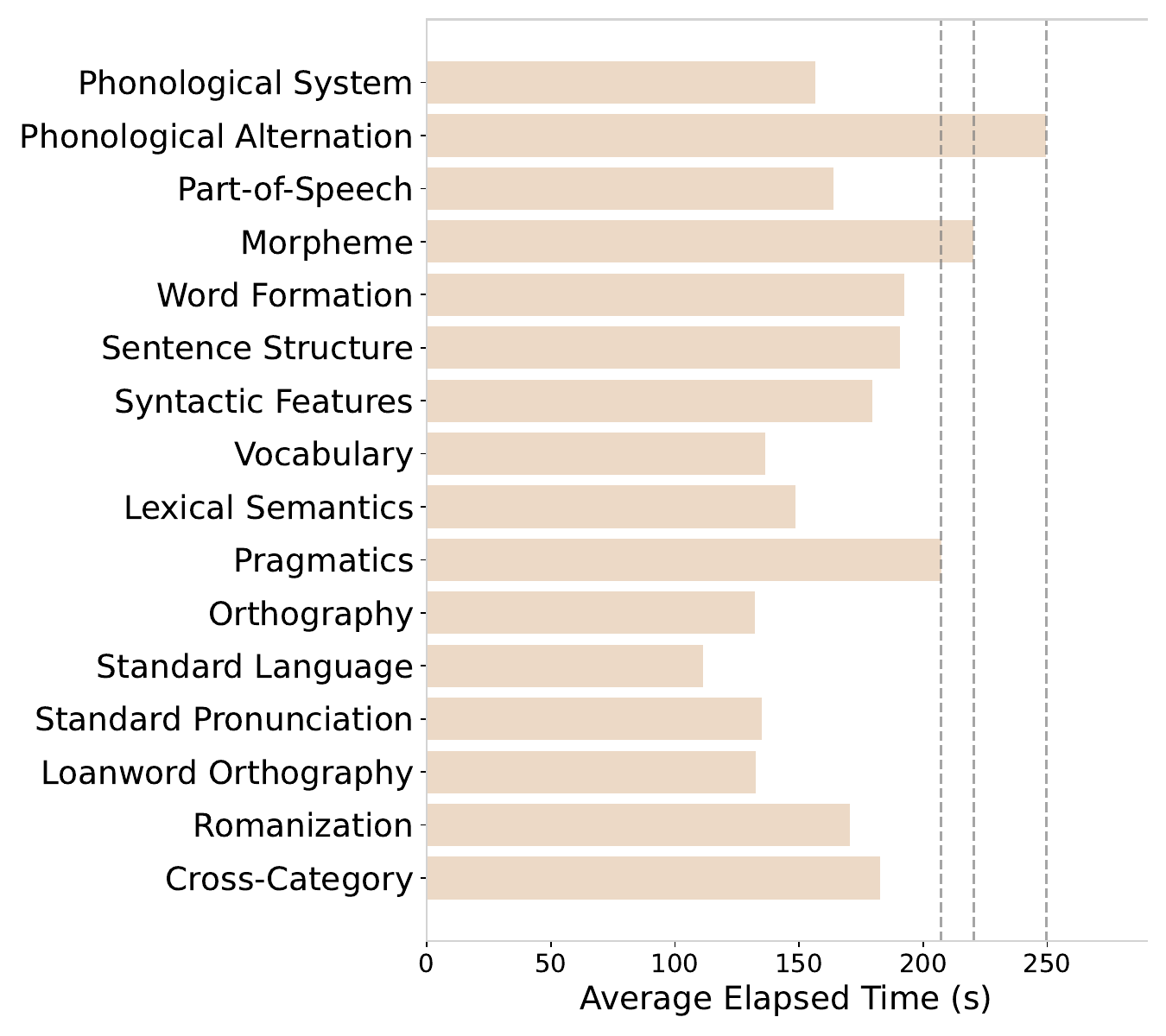}
    \caption{Comparison of the average \textit{thinking} time required to solve each question and the corresponding average score for each subcategory by the s1-32B model. Dashed lines indicate the three longest test times.}
    \label{fig:elapsed_time}
\end{figure}

We aimed to indirectly assess the level of difficulty that LLMs face across subcategories by measuring their thinking time. Concretely, we evaluated the thinking time of the s1-32B model across all 1,524 questions in KoGEM and calculated the average time per subcategory. 
As shown in Figure~\ref{fig:elapsed_time}, our results indicate that the model takes significantly longer to solve problems in \texttt{Phonological Alternation}, \texttt{Morpheme}, and \texttt{Pragmatics} subcategories. Notably, these subcategories also exhibit a relatively large performance gap compared to humans, as discussed in Section~\ref{analysis_subcategory}.
This suggests that, unlike humans, who leverage real-world experience and intuition to efficiently solve problems, LLMs struggle in these subcategories.

\subsection{Impact of Experiential Knowledge}
\label{additional_knowledge}
\paragraph{\textbf{Phonological Alternation}}
To examine the effects of experiential knowledge in \texttt{Phonological Alternation} subcategory, we consider human subvocalization—a phenomenon in which individuals mentally rehearse the pronunciation of a word upon seeing it \cite{working_memory-1974, smith1961psychology}. To obtain the pronunciation of the text, we use g2pK\footnote{\url{https://github.com/Kyubyong/g2pK}} to convert Korean graphemes into phonemes. For example, given the text ``오늘은 하늘이 맑습니다$_\text{ Today, the sky is clear}$'', g2pK generates ``오느른 하느리 막씀니다$_\text{ To-day, the skai iz kli-er}$''. We append this produced \textit{pronunciation text} to each \texttt{choice} in the prompt.
As shown in Figure~\ref{fig:exp_knowledge}(a), adding \textit{pronunciation text} improves performance by 3.1\% to 26.3\%. These findings demonstrate that incorporating pronunciation could significantly enhance the linguistic competence of LLMs in phonological tasks.
However, this improvement does not conclusively determine whether LLMs possess latent pronunciation knowledge that remains underutilized or simply lack such knowledge, suggesting the need for further analysis in future work.

\paragraph{\textbf{Morpheme}}
In morphological processing, humans intuitively decompose words into morphemes, deriving meaning from root words~\cite{berko1958child, taft1975lexical}. To test whether LLMs can similarly benefit from morpheme awareness, we use Kiwi\footnote{\url{https://github.com/bab2min/Kiwi}} to decompose words into morphemes and append this \textit{morpheme text} to each choice in the prompt. For instance, ``오늘은 하늘이 맑습니다$_\text{ Today, the sky is clear}$'' is segmented as \{오늘$_\text{today}$ / 은$_\text{(auxiliary particle)}$ / 하늘$_\text{sky}$ / 이$_\text{is}$ / 맑$_\text{clear}$ / 습니다$_\text{(present progressive ending)}$\}. As shown in Figure~\ref{fig:exp_knowledge}(b), adding \textit{morpheme text} improves performance by 7.1\% to 20.0\%. These results present that incorporating explicit morphological cues can significantly enhance the performance of LLMs in morphologically diverse tasks.
These significant improvements suggest that LLMs still have room for further improvement in understanding Korean morphological knowledge.

\section{Qualitative Evaluation of Generated Explanations}

\begin{table}[t!]
    \small
    \setlength{\tabcolsep}{1.7pt}             
    \renewcommand{\arraystretch}{1.50}       
    \begin{tabular}{lcccc}
        \toprule
        Model & Faithful & Coherence & Fluency & Relevance \\
        \midrule
        HC-HCX-003 & 0.80 & 0.86 & 0.98 & 0.92 \\
        Claude-3.5-Sonnet & 0.92 & 0.96 & 1.00 & 1.00 \\
        GPT-4o & 0.86 & 0.94 & 1.00 & 1.00 \\
        \midrule
        Total & 0.86 & 0.92 & 0.99 & 0.97\\
        \bottomrule
    \end{tabular}
    \caption{Qualitative analysis of the explanations generated by LLMs on the KoGEM Benchmark. Specifically, we evaluated the outputs of a top Korean-centric model, HyperClova-HCX-003 (for short HC-HCX-003), and two top English-centric models, excluding the test-time scaled model. Three Korean native speakers assessed the texts based on four criteria: Faithfulness, Coherence, Fluency, and Relevance. The evaluations produced a Fleiss' Kappa of 0.42, indicating moderate agreement.}
    \label{tab:qualitative_analysis}
\end{table}

We extended the evaluation beyond multiple-choice accuracy by prompting models to generate textual explanations for their choices and conducting a qualitative analysis of these responses. Inspired by \citet{fabbri-etal-2021-summeval, liu-etal-2023-g, elangovan-etal-2024-considers}, we adopted an evaluation framework with four metrics:
`Faithfulness (for short, Faithful)' is the factual accuracy of the generated statement. 
`Coherence' is the logical consistency within the sentence.
`Fluency' is the grammatical and linguistic naturalness of the sentence. 
`Relevance' is the degree of semantic alignment between the generated rationale and the given question.

Table~\ref{tab:qualitative_analysis} presents the averaged scores for each metric across three open-source models: HC-HCX-003, Claude-3.5-Sonnet, and GPT-4o. The results show that two English-centric models achieved perfect fluency and relevance (1.00), and very high scores in coherence (0.94 and 0.96). In contrast, HC-HCX-003 showed relatively lower performance, particularly in faithfulness (0.80) and coherence (0.86), despite maintaining strong fluency (0.98) and relevance (0.92).
Overall, the models consistently produced relevant and coherent justifications. However, variations in faithfulness suggest that some models may still generate explanations that are linguistically well-formed but not entirely aligned with factual content. These findings highlight the importance of evaluating generated explanations through a multifaceted lens.

\section{Related Work}

\paragraph{Linguistic Competence in NLP}
To evaluate the linguistic competence of language models, previous studies have employed probing methods~\cite{conneau-etal-2018-cram, hewitt-manning-2019-structural, tenney-etal-2019-bert, tenney2019what, pimentel-etal-2020-information}, which assess the extent to which linguistic information is encoded in the hidden representations of pretrained models. While most of these studies have primarily focused on morphological and syntactic phenomena in English, more recent work has expanded the scope of evaluation to encompass additional linguistic dimensions, including semantics, phonology, and pragmatics~\cite{
begus2025,
Anna2024dissociating, waldis-etal-2024-holmes}.

Although probing methods provide insights into the presence of linguistic information in model representations, they are limited in explicitly evaluating the systematic grammatical competence of LLMs. Moreover, there is a lack of systematic approaches to assessing the linguistic competence of language models in Korean. To this end, we propose a granular and comprehensive benchmark to systematically evaluate the linguistic knowledge of LLMs in Korean.

\paragraph{Korean Grammatical Knowledge Evaluation}
Several benchmarks have been developed to evaluate Korean linguistic knowledge. For example, \citet{son-etal-2024-hae} focused on lexical knowledge, such as identifying Korean equivalents for loanwords. \citet{kim-etal-2024-click} conducted a general evaluation of Korean grammar, but their work lacked clarity regarding the scope of grammatical knowledge covered. Additionally, while \citet{koo-et-al-K-NCT} and \citet{yoon-etal-2023-kagas} examined specific types of grammatical errors in Korean, they focused on morphosyntactic and orthographic correctness, leaving broader aspects of linguistic competence unexplored. In contrast to these previous works, KoGEM offers a comprehensive and systematic evaluation for the grammatical knowledge of LLMs in Korean. We further provide insights into the linguistic competence of LLMs in Korean by comparing their performance with that of humans.

\section{Conclusion}
In this paper, we illuminated the distinctive strengths and weaknesses of LLMs in linguistic competence compared to humans through the lens of the \underline{Ko}rean \underline{G}rammar \underline{E}valuation Bench\underline{M}ark, KoGEM.
We conducted extensive experiments on a wide range of LLMs and found that while they excel in tasks requiring straightforward memorization or syntactic rule application, they struggle with linguistic subcategories that demand intuitive reasoning and real-world knowledge.
In contrast, humans demonstrated relatively superior performance in these challenging areas for LLMs, leveraging intuitive understanding and experiential knowledge from real-world contexts.
Furthermore, our in-depth analysis revealed that integrating experiential knowledge can significantly enhance the linguistic competence of LLMs. These insights highlight the importance of targeted benchmarks like KoGEM for assessing and advancing the multifaceted linguistic competence of LLMs. By polishing every facet of linguistic competence in LLMs through fine-grained comparisons with humans, we provide valuable insights to advance LLMs and propel their journey toward genuine linguistic competence.


\section*{Limitations}
Although our study has demonstrated its value in assessing Korean grammatical competence in both LLMs and humans, there are several limitations that could benefit from further examination. First, while KoGEM was constructed based on a fine-grained taxonomy of grammar, it is currently confined to the Korean language. Expanding our approach to other languages requires incorporating the unique linguistic knowledge of each language. However, since the core linguistic categories (i.e., phonology, morphology, syntax, and semantics) are universally applicable, we believe that our experimental pipeline can serve as a foundation for further investigations across different languages.

Second, we categorized grammatical knowledge based on theoretical linguistics and standardized prescriptive grammar. However, since prescriptive grammar, which dictates how language should be used, focuses on the correct use of language, there could be a gap between prescriptive grammar and actual language use by humans in the real world. To help bridge this gap, incorporating a descriptive grammar, which describes how language is used, is expected to be necessary in future works.

Lastly, to ensure a more accurate and reliable evaluation, we addressed two potential concerns. One concern is that the observed advantage in definitional tasks could stem from data contamination between our benchmark and the LLMs’ training corpora. Given the lack of transparency in LLM training data, we cannot completely exclude the possibility of data contamination. To address this concern, we evaluated 27 diverse models to capture general trends across architectures and datasets. The consistent performance patterns suggest that the outcome is unlikely to result solely from data memorization.
Another factor is prompt design, which can affect model behavior. As an initial design choice, we aligned the prompt with that presented to human participants to ensure consistency. Future work could explore whether alternative designs yield greater robustness or deeper insights.

\section*{Acknowledgments}
This work was supported by the National Research Foundation of Korea (NRF) grant funded by the Korea government (MSIT) (No.RS-2025-00517221 and No.RS-2024-00415812) and Institute of Information \& communications Technology Planning \& Evaluation (IITP) grant funded by the Korea government (MSIT) (No.RS-2024-00439328, Karma: Towards Knowledge Augmentation for Complex Reasoning (SW Starlab), No.RS-2024-00457882, AI Research Hub Project, and No.RS-2019-II190079, Artificial Intelligence Graduate School Program (Korea University)).
This work was additionally supported by the 2024 BK21 Four Project ``Research Challenger Program for Graduate Students in Interdisciplinary Research Groups'' at the Center for Teaching and Learning Support, Korea University. The work of the author, Nayeon Kim, was also supported by Basic Science Research Program through the National Research Foundation of Korea (NRF) funded by the Ministry of Education (No.RS-2023-00271662). 
Lastly, We especially thank Youngje Lee, Associate Professor in the Department of Korean Language and Literature in Korea University, for his insightful comments and suggestions.

\bibliography{custom}

\clearpage

\appendix

\section{Details of LLM Baselines}
\label{llm_details}
In this paper, we use the eight Korean-centric LLMs primarily trained on Korean data and the eleven English-centric LLMs primarily trained on English data for zero-shot evaluation on KoGEM as follows:

\begin{itemize}[left=0cm]
    \item \textbf{Korean-centric LLMs}
    \begin{enumerate}[left=0cm]
        \item Bllossom-8B~\cite{choi-etal-2024-optimizing}: 
        This is a Korean-English bilingual language model based on the open-source LLama3. It enhances the connection of knowledge between Korean and English.
        \item SOLAR-v1.0-10.7B-Instruct~\cite{kim-etal-2024-solar}: 
        This is an advanced LLM with 10.7 billion parameters. It is trained by utilizing instruction fine-tuning methods, including supervised fine-tuning (SFT) and direct preference optimization (DPO)~\cite{Rafailov-DPO-2023}.
        \item KULLM-3-10.7B~\cite{kullm3}: This model is instruction-tuned from the \texttt{upstage/SOLAR-10.7B-v1.0}~\cite{kim-etal-2024-solar} model. This model is trained using the data, such as \citet{peng2023instruction} and mixed Korean instruction data (gpt-generated, hand-crafted, etc).
        \item EEVE-v1.0-10.8B-Instruct~\cite{kim2024efficient}: 
        This model is a fine-tuned version of \texttt{yanolja/EEVE-Korean-10.8B-v1.0},\footnote{\url{https://huggingface.co/yanolja/EEVE-Korean-10.8B-v1.0}} which is a Korean vocabulary-extended version of \texttt{upstage/SOLAR-10.7B-v1.0}~\cite{kim-etal-2024-solar}. Specifically, this model is trained by utilizing DPO.\footnote{\url{https://github.com/axolotl-ai-cloud/axolotl}}
        \item EXAONE-3.5-7.8B-Instruct~\cite{exaone-3.5}: 
        This model is an instruction-tuned bilingual (English and Korean) generative model, developed and released by LG AI Research. This model is trained to utilize the system prompt.\footnote{\url{https://huggingface.co/LGAI-EXAONE/EXAONE-3.5-7.8B-Instruct}}
        \item EXAONE-3.5-32B-Instruct~\cite{exaone-3.5}: 
        This model shares the same architecture as above \texttt{EXAONE-3.5-7.8B-Instruct}, differing only in size.
        \item HyperCLOVA-HCX-DASH-001~\cite{yoo2024hyperclovaxtechnicalreport}: 
        This is an optimized version of \texttt{HyperCLOVA-HCX-003} that offers faster response times and cost efficiency, making it suitable for simpler tasks while maintaining robust performance.
        \item HyperCLOVA-HCX-003~\cite{yoo2024hyperclovaxtechnicalreport}: 
        This is a foundational model in NAVER's HyperCLOVA X suite, designed for complex and sophisticated tasks, delivering high-quality responses.
    \end{enumerate}
    
    \item \textbf{English-centric LLMs}
    \begin{enumerate}[left=0cm]
        \item Gemma-2-9B-Instruct~\cite{team2024gemma}: 
        Developed by Google, this model is part of the Gemma series and contains 9 billion parameters. It is a text-to-text, decoder-only large language model, with open weights for both pre-trained and instruction-tuned variants. Gemma models are well-suited for a variety of text generation tasks, including question answering, summarization, and reasoning.
        \item Gemma-2-27B-Instruct~\cite{team2024gemma}: 
        This model is the largest in Google's Gemma series, featuring 27 billion parameters. It is designed to deliver high performance across various natural language processing tasks, benefiting from its extensive parameter count and advanced training methodologies.
        \item Qwen2.5-7B-Instruct~\cite{qwen2.5}: 
        Developed by Alibaba, this model is part of the Qwen2.5 series and contains 7 billion parameters. It is designed to handle various natural language understanding and generation tasks, supporting multiple languages, including English and Chinese.
        \item Qwen2.5-14B-Instruct~\cite{qwen2.5}: 
        This model is a mid-sized variant in Alibaba's Qwen2.5 series, featuring 14 billion parameters. It offers enhanced performance in language understanding and generation tasks, with support for multiple languages and a context length of up to 128,000 tokens.
        \item Qwen2.5-32B-Instruct~\cite{qwen2.5}: 
        This model is part of Alibaba's Qwen2.5 series, featuring 32 billion parameters. It supports a context length of up to 128,000 tokens and is designed to handle complex tasks across multiple languages, including English and Chinese. All Qwen 2.5 series are open-source under the Apache 2.0 license.
        \item DeepSeek-R1-Distill-Qwen-14B~\cite{liu2024deepseek}: 
        Developed by DeepSeek, this model is a distilled version of their R1 model, based on Qwen2.5-14B, containing 14 billion parameters. It has been fine-tuned using reasoning data generated by DeepSeek-R1, resulting in enhanced performance in reasoning tasks.
        \item DeepSeek-R1-Distill-Qwen-32B~\cite{liu2024deepseek}: 
        This model is another distilled variant from DeepSeek, based on Qwen2.5-32B, featuring 32 billion parameters. It has been fine-tuned with reasoning data from DeepSeek-R1, achieving state-of-the-art results in various benchmarks.
        \item s1-32B~\cite{s1-32b-stanford-2025}: 
        Developed by Stanford and the University of Washington, s1-32B is a fine-tuned version of Qwen2.5-32B-Instruct. It was optimized for reasoning tasks using 1,000 high-quality samples. The model employs a novel "budget forcing" technique to enhance reasoning efficiency and outperforms OpenAI's o1-preview on certain benchmarks.
        \item Llama-3.1-8B-Instruct~\cite{grattafiori2024llama3herdmodels}: 
        Developed by Meta AI, this model is part of the Llama-3.1 series and contains 8 billion parameters. It has been instruction-tuned to enhance its performance in various natural language understanding and generation tasks.
        \item Llama-3-70B~\cite{grattafiori2024llama3herdmodels}: 
        Developed by Meta AI, Llama-3-70B is a large language model with 70 billion parameters. It has been pre-trained on approximately 15 trillion tokens from publicly available sources. The model is designed to be multilingual and multimodal, with enhanced capabilities in coding and reasoning.
        \item Llama-3.1-405B~\cite{grattafiori2024llama3herdmodels}: 
        This is an expanded version of the Llama series, featuring 405 billion parameters. It demonstrates superior performance in general knowledge and reasoning tasks, achieving high scores on benchmarks such as MMLU-Pro and MMLU-redux.
        \item Gemini-1.5-flash~\cite{geminiteam2024gemini15unlockingmultimodal}: 
        Developed by Google DeepMind, Gemini-1.5-flash is a multimodal language model capable of processing text, images, audio, and video. It is designed for real-time interactions and has been integrated into various Google products, including Bard and Pixel smartphones.
        \item Gemini-2.0-flash-exp~\cite{geminiteam2024gemini15unlockingmultimodal}: 
        An experimental update to the Gemini series, this model offers improved speed and performance over its predecessors. It introduces features such as a Multimodal Live API for real-time audio and video interactions, enhanced spatial understanding, and integrated tool use, including Google Search.
        \item Claude-3-haiku~\cite{claude3}: 
        Developed by Anthropic, Claude-3-haiku is a large language model designed for complex conversational tasks. It features a context window of up to 200,000 tokens, allowing it to process extensive text sequences effectively.
        \item Claude-3.5-Sonnet~\cite{claude3.5}: 
        An enhanced version of the Claude series, this model offers improved performance in language understanding and generation tasks. It maintains a large context window and has been fine-tuned for better alignment with human preferences.
        \item GPT-3.5-turbo~\cite{openai2023gpt35turbo}: 
        Developed by OpenAI, GPT-3.5-turbo is an improvement over the original GPT-3.5 model, offering better accuracy in responses. It has been widely used in applications requiring natural language understanding and generation. 
        \item GPT-4o-mini~\cite{openai2024gpt4technicalreport}: 
        A smaller and more cost-effective version of OpenAI's GPT-4o, this model is capable of processing text, images, and audio. It offers rapid response times and has been integrated into various applications for real-time interactions.
        \item GPT-4o~\cite{openai2024gpt4technicalreport}: 
        OpenAI's GPT-4o is a multimodal model capable of analyzing and generating text, images, and sound. It exhibits rapid response times comparable to human reactions and has enhanced performance in non-English languages.
        \item o1-preview~\cite{o1_report}: 
        Introduced by OpenAI, o1-preview is designed to solve complex problems by spending more time ``thinking'' before responding. It outperforms previous models in areas like competitive programming, mathematics, and scientific reasoning.
    \end{enumerate}
\end{itemize}

\section{Details of LLM Evaluation}
\label{appendix: details_of_llm_evaluation}
In this section, we explain the details of LLM evaluation settings, such as device settings, hyperparameters, and prompt design, as the input for a single run of zero-shot evaluation.

\subsection{Device Settings}
We use the Hugging Face library\footnote{\url{https://huggingface.co/models}} with local GPU settings for open-source models, including all Korean-centric LLMs except for the HyperCLOVA X series, the Gemma series, the Qwen2.5 series, the DeepSeek-R1-Distill series, and the Llama-3.1-8B model among English-centric LLMs. For models larger than 20B parameters, we utilize two RTX A6000 GPUs, while for relatively smaller models, we use a single RTX A6000 GPU. For closed-source models, we use the corresponding API provided for each model.

\subsection{Hyperparameters}
To ensure reproducibility, we set the temperature to 0 or very close to 0 (e.g., 1e-10). However, for GPT-3.5-turbo and o1-preview, we cannot customize the temperature setting. This is because the GPT-3.5-turbo model does not support temperature adjustment~\cite{openai2023gpt35turbo}, and the function is unavailable in o1-preview. Additionally, we fix the random seed to 42 across all experiments.

To manage API usage costs, we limit the maximum number of output tokens to 200. Despite this restriction, we confirm that no outputs are truncated, as the input prompts explicitly set a maximum output length of 100 characters.

\subsection{Prompt Designs for Main Results}
\label{prompt_designs}
In this section, we describe the prompts used in our main experiments. The prompts are categorized into four types (T1, T2, T3, and T4) based on the given information, which may include {\texttt{passage}, \texttt{question}, \texttt{paragraph}, \texttt{choices}}. Additionally, to facilitate understanding, we provide the English translation of the full T1 prompt at the end of the examples below.

\subsection{Statistics of Prompt Lengths}
\begin{figure}
    \centering
    \includegraphics[width=1.0\linewidth]{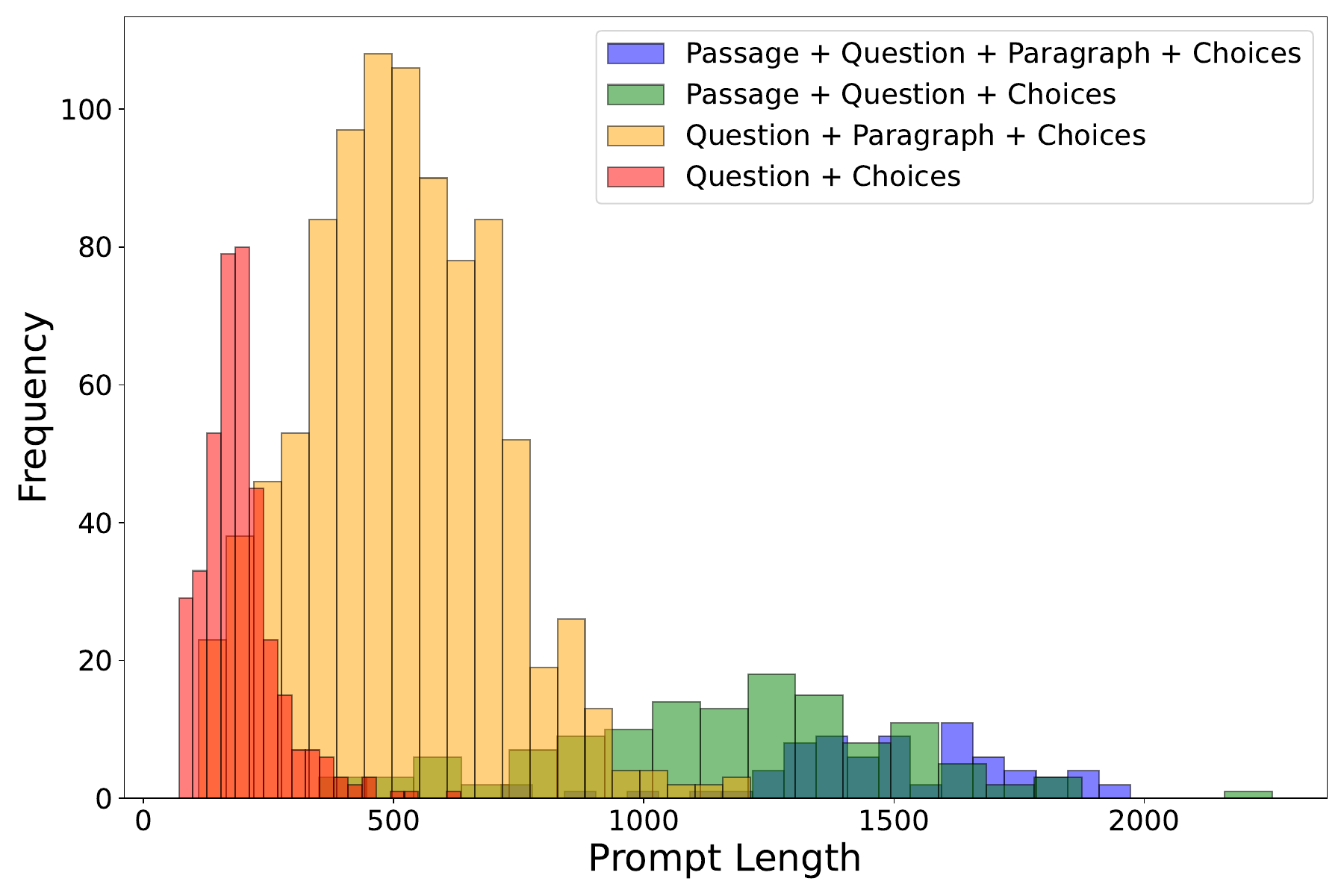}
    \caption{Distribution of the KoGEM benchmark by prompt type. These four types correspond to T1–T4 described earlier in Section~\ref{prompt_designs}.}
    \label{fig:prompt_lenghts}
\end{figure}

Figure~\ref{fig:prompt_lenghts} summarizes the average lengths and instance counts by prompt type. On average, each prompt, including (passage, question, paragraph, and choices), contains 534 characters, with many exceeding 1,000 characters due to full-passage inclusion. This structure effectively evaluates prompts of varying lengths, requiring integration of broad linguistic context.
\vspace{0.5cm}

\setlength{\fboxsep}{5pt} 

\begin{itemize}[left=0cm]
\item \textbf{T1} \{\texttt{question}+\texttt{choices}\}\\
    \noindent
    \fbox{
        \begin{minipage}{6.8cm} \small
            [system] \\
            다음은 한국어 언어 이해에 대한 객관식 문제입니다. 주어진 질문에 대한 정답으로 올바른 번호를 선택지에서 고르고, 그에 맞는 해설을 100자 내로 설명하시오.\\
            
            [user] \\
            질문: 다음 선택지 1 부터 \{4 or 5\} 중 \{\texttt{question}\} \\
            선택지: \{\texttt{choices}\}
      \end{minipage}
    }
    
    \item \textbf{T2} \{\texttt{question}+\texttt{paragraph}+\texttt{choices}\}\\
    \noindent
    \fbox{
        \begin{minipage}{6.8cm} \small
            [system] \\
            다음은 한국어 언어 이해에 대한 객관식 문제입니다. 주어진 설명을 보고, 질문에 대한 정답으로 올바른 번호를 선택지에서 고르고, 그에 맞는 해설을 100자 내로 설명하시오.\\
            
            [user] \\
            설명: \{\texttt{paragraph}\} \\
            질문: 다음 선택지 1 부터 \{4 or 5\} 중 \{\texttt{question}\} \\
            선택지: \{\texttt{choices}\}
      \end{minipage}
    }
    
    \item \textbf{T3} \{\texttt{passage}+\texttt{question}+\texttt{choices}\}\\
    \noindent
    \fbox{
        \begin{minipage}{6.8cm} \small
            [system] \\
            다음은 한국어 언어 이해에 대한 객관식 문제입니다. 주어진 지문을 보고, 질문에 대한 정답으로 올바른 번호를 선택지에서 고르고, 그에 맞는 해설을 100자 내로 설명하시오.\\
            
            [user] \\
            지문: \{\texttt{passage}\} \\
            질문: 다음 선택지 1 부터 \{4 or 5\} 중 \{\texttt{question}\} \\
            선택지: \{\texttt{choices}\}
      \end{minipage}
    }

    \item \textbf{T4} \{\texttt{passage}+\texttt{question}+\texttt{paragraph}+\texttt{choices}\}\\
    \noindent
    \fbox{
        \begin{minipage}{6.8cm} \small
            [system] \\
            다음은 한국어 언어 이해에 대한 객관식 문제입니다. 주어진 지문과 설명을 보고, 질문에 대한 정답으로 올바른 번호를 선택지에서 고르고, 그에 맞는 해설을 100자 내로 설명하시오.\\
            
            [user] \\
            지문: \{\texttt{passage}\} \\
            설명: \{\texttt{paragraph}\} \\
            질문: 다음 선택지 1 부터 \{4 or 5\} 중 \{\texttt{question}\} \\
            선택지: \{\texttt{choices}\}
      \end{minipage}
    }

    \item Translation of \textbf{T4}\\
    \noindent
    \fbox{
        \begin{minipage}{6.8cm} \small
            [system] \\
            The following is a multiple-choice question about Korean language comprehension. Based on the given passage and explanation, choose the correct number from the choices as the answer to the question and provide a corresponding explanation within 100 characters.\\
            
            [user] \\
            Passage: \{\texttt{passage}\} \\
            Paragraph: \{\texttt{paragraph}\} \\
            Question: Choose from options 1 to \{4 or 5\} for \{\texttt{question}\} \\
            Choices: \{\texttt{choices}\}
      \end{minipage}
    }
\end{itemize}

\subsection{Prompt Design for Addition of Pronunciation Experiment}
In Section~\ref{phonology_analysis}, we conducted additional experiments on the phonological alternation task by incorporating pronunciation information into the choices within the prompt. Figure~\ref{fig:example_phonology} represents a typical case from this experiment.

\section{Details of Crowdsourcing for Human Evaluation}
\label{appendix: human_evaluation}
In this section, we provide details about the crowdsourcing process used to evaluate human performance. As mentioned in Section~\ref{human_evaluation}, we assess human performance across five exams: HSQE, LCSE (G9 and G7), and NCSE (G9 and G7), using crowdsourcing. Figure~\ref{fig:human_instruction} presents the instructions provided to crowdworkers, while Figure~\ref{fig:human_test_sample} shows an actual test sample given to participants. We collected over 10 responses per question from a total of 352 participants across 583 questions. 
To maintain concentration and prevent crowdworker bias, we assign a maximum of 20 questions per crowdworker.
The cost per response was approximately \$0.24.

\begin{figure}
    \centering
    \includegraphics[width=1.0\linewidth]{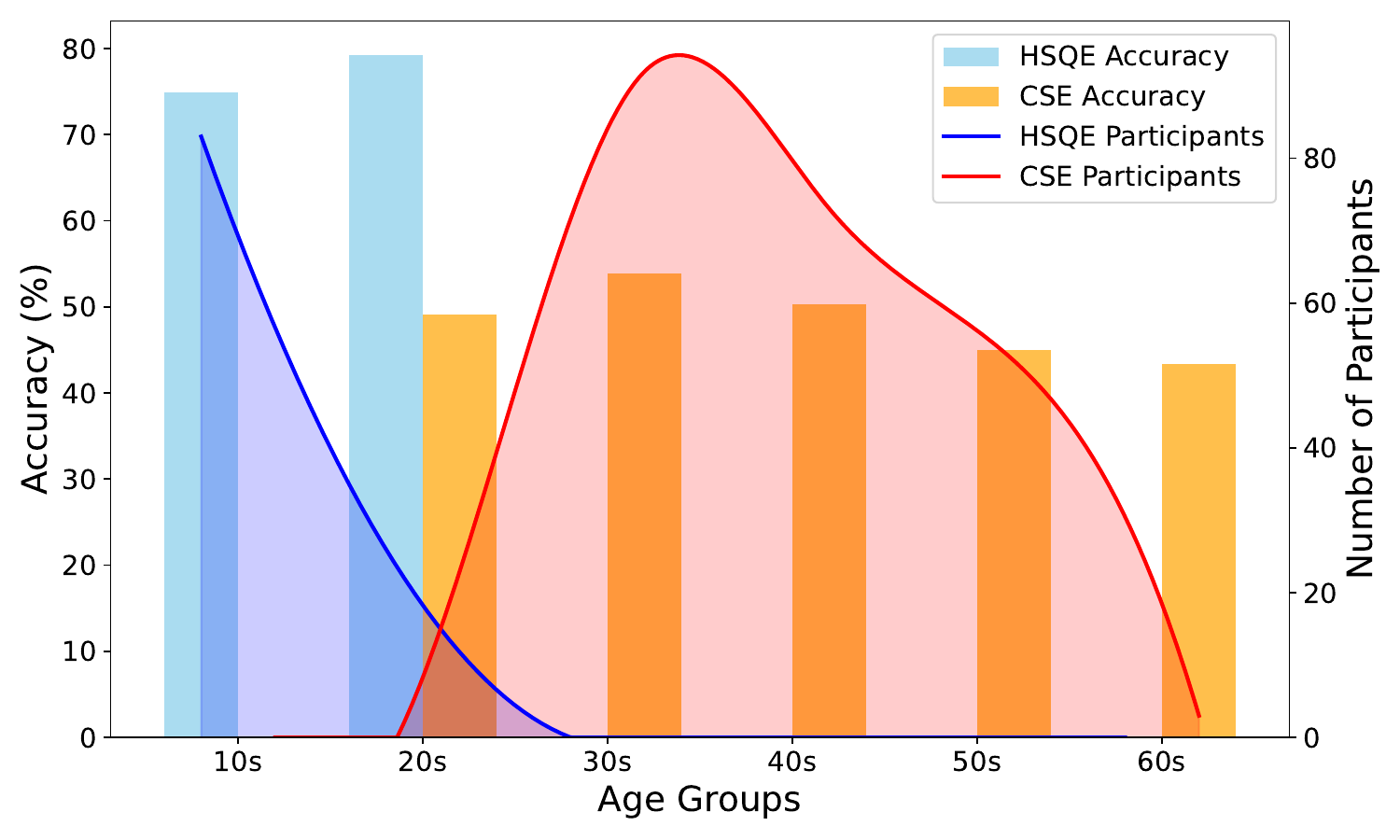}
    \caption{Age distribution and accuracy of participants in HSQE and CSE tests. The line graph illustrates the number of participants across different age groups, while the bar graph represents their corresponding accuracy rates.}
    \label{fig:human_statistics}
\end{figure}

For the CSE, responses were collected from a diverse age group of active civil servants, ranging from their teens to their 60s, to mitigate potential biases associated with crowdworkers. In contrast, the HSQE focused on first-year university students who had completed the relevant curriculum within the past year. The age distribution of participants and their accuracy across age groups are presented in Figure~\ref{fig:human_statistics}. Additionally, to minimize gender bias, we made efforts to balance the gender distribution as evenly as possible.

\definecolor{skyblue}{RGB}{47, 170, 218}

\begin{figure*}[t!]
    \centering
    \includegraphics[width=1.0\linewidth]{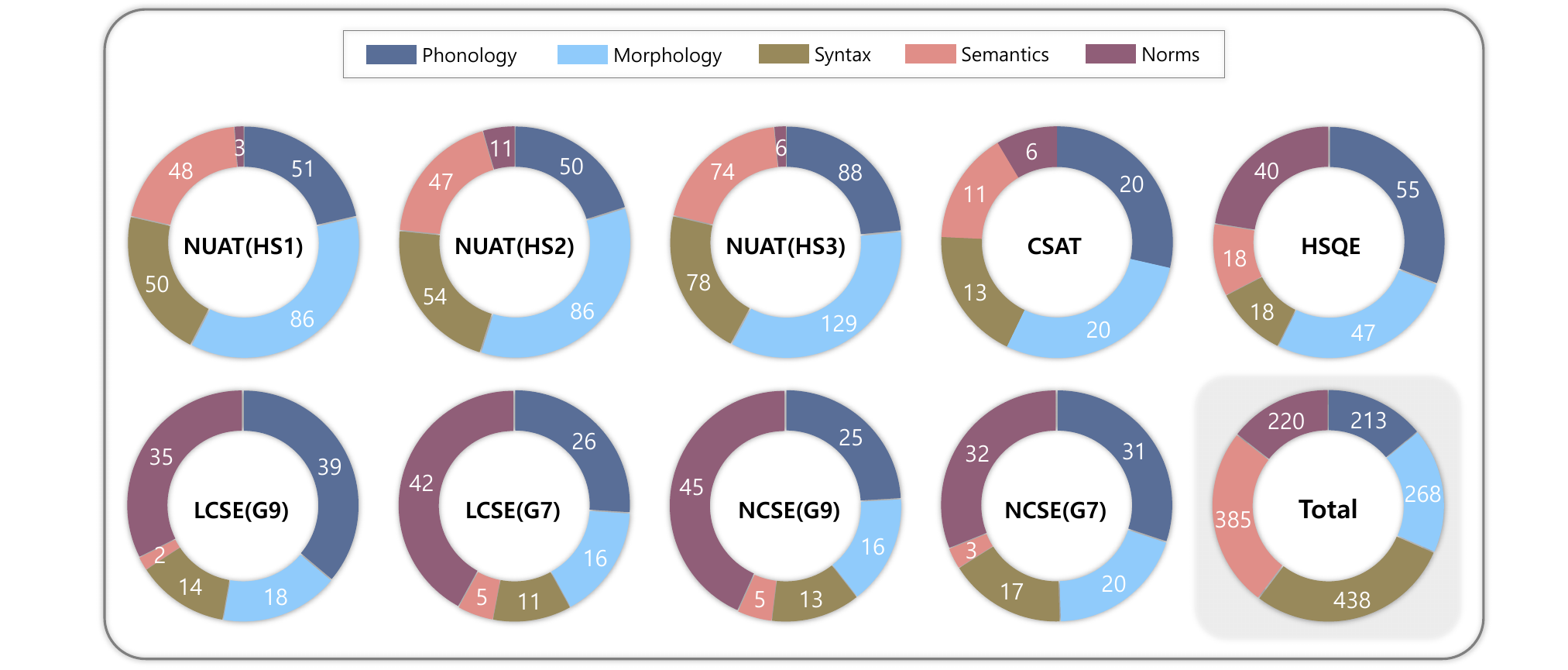}
    \caption{The specific numbers and ratios of linguistic categories and subcategories for each source exam.}
    \label{fig:subcategory_numbers}
\end{figure*}

\section{Details of KoGEM}
In this section, we provide detailed descriptions of KoGEM, including source data, preprocessing rules, and subcategories of Korean grammar taxonomy.

\subsection{Source Data}
\label{sec:appendix_source_data}
We detail the characteristics of each source data in KoGEM, and their copyright and license. 

\subsubsection*{Detailed Description}
\begin{itemize}[left=0.0cm]
\item College Scholastic Ability Test (CSAT) is an exam administered by the Korea Institute for Curriculum and Evaluation to select qualified individuals for university admission. CSAT is administered once a year.\footnote{\url{https://www.suneung.re.kr}}
    
\item National United Achievement Test (NUAT) is a mock exam conducted in a format similar to the CSAT. NUAT is administered by the Seoul Education Research and Information Institute and is regularly taken by high school students across all grades. NUAT is conducted four times a year for first- and second-year high school students, and six times a year for third-year students.\footnote{\url{https://www.jinhak.or.kr}}
    
\item High School Qualification Exam (HSQE) is an exam administered by the Korea Institute for Curriculum and Evaluation to assess the qualifications required for high school graduation. We solely use the Korean language section of the High School Qualification Exam. HSQE is held twice a year.\footnote{\url{https://www.kice.re.kr}}
    
\item Civil Service Exam (CSE) is an exam administered by the Ministry of Personnel Management to recruit national and local civil servants. We extract grammar questions from the Korean language section of the national and local 7th- and 9th-grade recruitment exams. Although there are more types of CSEs, we use exams for which copyright issues have been resolved. CSE is conducted once a year.\footnote{\url{https://www.gosi.kr}}

\end{itemize}

\subsubsection*{License}
The institutions providing our source exams have made these works available under the ``Public Copyright Free Use Permit Standard (Korean Open Government License, KOGL) Type 1.'' Under KOGL Type 1, users can use public works freely and without fee, regardless of their commercial use, and can change or modify to create secondary works.\footnote{\url{https://www.kogl.or.kr/info/license.do}} Additionally, we have contacted each institution and obtained permission for research purposes.

\newtcbox{\inlinebox}[1][]{
  colback=gray!10,    
  colframe=white,     
  boxrule=0pt,        
  arc=3pt,            
  left=2pt,           
  right=2pt,          
  top=1pt,            
  bottom=1pt,         
  on line,            
  #1
}

\subsection{Data Preprocessing}
\label{sec:appendix_preprocessing}
Since we extract textual data from exam images using HyperCLOVA X OCR~\footnote{\url{https://clova.ai/hyperclova}}, the extracted data often contain numerous errors, such as typos, fragmented characters, and unstructured or inconsistent formats. To standardize the structure and representation of the data, we utilize HTML formatting. Below are the formatting rules we follow:
\begin{itemize}
    \item \textbf{Underline.} Underline: We enclose underlined text within the \inlinebox{\texttt{<u>}} tag. If the underlined text includes a specific symbol, such as ㉠ we wrap the text using the corresponding symbol tag, e.g., \inlinebox{\texttt{<㉠>}}.
    \item \textbf{Boldface.} Text requiring emphasis is enclosed within the \inlinebox{\texttt{<b>}} tag for bold formatting.
    \item \textbf{Box.} Text to be highlighted within a box is wrapped using the \inlinebox{\texttt{<box>}} tag.
    \item \textbf{Section.} Text belonging to the same section is grouped using a designated section tag. Examples include \inlinebox{\texttt{<가>}}, \inlinebox{\texttt{<A>}}, \inlinebox{\texttt{<예>}}, \inlinebox{\texttt{<보기>}}, etc.
    \item \textbf{Table.} Rows are separated by the newline character (\textbackslash n). Within a row, the header column is distinguished by a colon (:), and other columns are separated using a slash (/).
    \item \textbf{Abbreviated Expression.} When more than three dots appear in the text, we standardize them to three unified dots (···).
    \item \textbf{Bullet Point.} We preserve the original bullet points described in the image whenever possible. However, for symbols with similar appearances (e.g., hollow middle circles or lower hollow circles), we assign a single representative symbol for uniformity.
    \item \textbf{Text with Circle.} If a piece of text is circled, we represent it by enclosing it in parentheses.
    \item \textbf{Text with Rectangular.} If text is enclosed within a rectangle, we represent it using square brackets.
\end{itemize}

\begin{table*}[t!]
    \small
    \setlength{\tabcolsep}{3.4pt}       
    \renewcommand{\arraystretch}{1.1}    
    \begin{tabular}{clcccccccccc}
        \toprule
        \multicolumn{1}{c|}{Language} & \multicolumn{1}{c|}{Model} & \begin{tabular}[c]{@{}c@{}}NUAT\\ (HS1)\end{tabular} & \begin{tabular}[c]{@{}c@{}}NUAT\\ (HS2)\end{tabular} & \begin{tabular}[c]{@{}c@{}}NUAT\\ (HS3)\end{tabular} & CSAT & HSQE & \begin{tabular}[c]{@{}c@{}}LCSE\\ (G9)\end{tabular} & \begin{tabular}[c]{@{}c@{}}LCSE\\ (G7)\end{tabular} & \begin{tabular}[c]{@{}c@{}}NCSE\\ (G9)\end{tabular} & \begin{tabular}[c]{@{}c@{}}NCSE\\ (G7)\end{tabular} & Avg. \\ 
        \midrule
        \multicolumn{2}{c|}{Random} & 20.00 & 20.00 & 20.00 & 20.00 & 25.00 & 25.00 & 25.00 & 25.00 & 25.00 & 21.95 \\
        \midrule
        \multicolumn{1}{c|}{\multirow{8}{*}{Korean}} & \multicolumn{1}{l|}{Bllossom-8B} & 27.31 & 24.60 & 25.07 & 20.00 & 35.96 & 25.93 & 29.00 & 31.73 & 24.27 & 27.10 \\
        \multicolumn{1}{c|}{} & \multicolumn{1}{l|}{SOLAR-v1.0-10.7B-Instruct} & 33.61 & 24.19 & 24.80 & 22.86 & 33.71 & 19.44 & 29.00 & 27.88 & 28.16 & 27.36 \\
        \multicolumn{1}{c|}{} & \multicolumn{1}{l|}{KULLM-3-10.7B} & 26.89 & 26.21 & 21.07 & 25.71 & 32.58 & 32.41 & 24.00 & 31.73 & 29.13 & 26.64 \\ 
        \multicolumn{1}{c|}{} & \multicolumn{1}{l|}{EEVE-v1.0-10.8B-Instruct} & 31.09 & 24.60 & 25.60 & 22.86 & 44.38 & 36.11 & 39.00 & 28.85 & 26.21 & 30.25 \\
        \multicolumn{1}{c|}{} & \multicolumn{1}{l|}{EXAONE-3.5-7.8B-Instruct} & 30.25 & 27.02 & 30.93 & 28.57 & 47.19 & 43.52 & 38.00 & 35.58 & 30.10 & 33.60 \\
        \multicolumn{1}{c|}{} & \multicolumn{1}{l|}{EXAONE-3.5-32B-Instruct} & 40.76 & 32.26 & 33.07 & 28.57 & 57.87 & 41.67 & 44.00 & 41.35 & 36.89 & 38.98 \\
        \multicolumn{1}{c|}{} & \multicolumn{1}{l|}{HyperCLOVA-HCX-DASH-001} & 29.41 & 21.37 & 26.40 & 27.14 & 46.07 & 36.11 & 29.00 & 40.38 & 34.95 & 30.77 \\
        \multicolumn{1}{c|}{} & \multicolumn{1}{l|}{HyperCLOVA-HCX-003} & 48.74 & 38.31 & 35.73 & 27.14 & 64.61 & 49.07 & 46.00 & 56.73 & 41.75 & 44.62 \\
        \midrule
        \multicolumn{1}{c|}{\multirow{18}{*}{English}} & \multicolumn{1}{l|}{Gemma-2-9B-Instruct} & 34.87 & 27.42 & 30.40 & 21.43 & 49.44 & 36.11 & 32.00 & 26.92 & 30.10 & 32.68 \\
        \multicolumn{1}{c|}{} & \multicolumn{1}{l|}{Gemma-2-27B-Instruct} & 39.50 & 29.03 & 35.20 & 20.00 & 54.49 & 30.56 & 31.00 & 38.46 & 30.10 & 35.70 \\
        \multicolumn{1}{c|}{} & \multicolumn{1}{l|}{Qwen2.5-7B-Instruct} & 37.82 & 32.26 & 28.27 & 27.14 & 47.19 & 32.41 & 26.00 & 29.81 & 34.95 & 33.27 \\
        \multicolumn{1}{c|}{} & \multicolumn{1}{l|}{Qwen2.5-14B-Instruct} & 46.22 & 37.50 & 34.40 & 31.43 & 60.67 & 30.56 & 38.00 & 43.27 & 33.98 & 40.22 \\
        \multicolumn{1}{c|}{} & \multicolumn{1}{l|}{Qwen2.5-32B-Instruct} & 48.32 & 41.53 & 39.73 & 37.14 & 58.43 & 45.37 & 37.00 & 39.42 & 31.07 & 43.04 \\
        \multicolumn{1}{c|}{} & \multicolumn{1}{l|}{DeepSeek-R1-Distill-Qwen-14B} & 45.80 & 34.27 & 34.13 & 20.00 & 55.62 & 37.04 & 32.00 & 33.65 & 32.04 & 37.73 \\
        \multicolumn{1}{c|}{} & \multicolumn{1}{l|}{DeepSeek-R1-Distill-Qwen-32B} & 50.42 & 42.34 & 42.67 & 35.71 & 65.17 & 41.67 & 35.00 & 35.58 & 34.95 & 44.55 \\
        \multicolumn{1}{c|}{} & \multicolumn{1}{l|}{s1-32B} & 52.94 & 48.39 & 43.73 & 47.14 & 67.98 & 42.59 & 37.00 & 47.12 & 34.95 & 48.03 \\
        \multicolumn{1}{c|}{} & \multicolumn{1}{l|}{Llama-3.1-8B-Instruct} & 30.67 & 25.40 & 21.60 & 31.43 & 37.08 & 24.07 & 31.00 & 29.81 & 27.18 & 27.62 \\
        \multicolumn{1}{c|}{} & \multicolumn{1}{l|}{Llama-3-70B} & 39.50 & 29.84 & 29.87 & 30.00 & 49.44 & 35.19 & 36.00 & 28.85 & 33.01 & 34.58 \\
        \multicolumn{1}{c|}{} & \multicolumn{1}{l|}{Llama-3.1-405B} & 57.98 & 43.55 & 46.93 & 48.57 & 60.67 & 37.96 & 38.00 & 39.42 & 33.98 & 47.18 \\
        \multicolumn{1}{c|}{} & \multicolumn{1}{l|}{Gemini-1.5-flash} & 49.58 & 43.95 & 43.20 & 25.71 & 65.73 & 39.81 & 47.00 & 42.31 & 32.04 & 45.34 \\
        \multicolumn{1}{c|}{} & \multicolumn{1}{l|}{Gemini-2.0-flash-exp} & 63.45 & 49.19 & 56.00 & 51.43 & 76.40 & 56.48 & 43.00 & 52.88 & 38.83 & 56.04 \\
        \multicolumn{1}{c|}{} & \multicolumn{1}{l|}{Claude-3-haiku} & 39.92 & 27.42 & 28.27 & 31.43 & 56.74 & 34.26 & 35.00 & 31.73 & 34.95 & 34.97 \\
        \multicolumn{1}{c|}{} & \multicolumn{1}{l|}{Claude-3.5-Sonnet} & {\ul 66.81} & {\ul 60.48} & {\ul 60.53} & {\ul 52.86} & 75.28 & 52.78 & 45.00 & 52.88 & 48.54 & {\ul 59.97} \\
        \multicolumn{1}{c|}{} & \multicolumn{1}{l|}{GPT-3.5-turbo} & 25.63 & 26.61 & 22.93 & 24.29 & 40.45 & 30.56 & 31.00 & 25.00 & 23.30 & 27.30 \\
        \multicolumn{1}{c|}{} & \multicolumn{1}{l|}{GPT-4o-mini} & 44.12 & 32.26 & 33.87 & 37.14 & 67.98 & 38.89 & 39.00 & 39.42 & 27.18 & 39.96 \\
        \multicolumn{1}{c|}{} & \multicolumn{1}{l|}{GPT-4o} & 63.45 & 54.84 & 53.87 & 45.71 & {\ul 79.21} & {\ul 55.56} & {\ul 51.00} & {\ul 55.77} & {\ul 49.51} & 57.87 \\
        \multicolumn{1}{c|}{} & \multicolumn{1}{l|}{o1-preview} & \textbf{86.97} & \textbf{83.87} & \textbf{79.47} & \textbf{75.71} & \textbf{94.38} & \textbf{78.70} & \textbf{76.00} & \textbf{74.04} & \textbf{61.17} & \textbf{81.04} \\
        \midrule
        \multicolumn{2}{c|}{LLMs Avg.} & 44.15 & 36.62 & 36.58 & 33.23 & 56.47 & 39.44 & 37.70 & 39.28 & 34.20 & 40.24 \\
        \midrule
        \multicolumn{2}{c|}{Human} & 72.66 & 61.58 & 72.45 & 65.88 & 78.60 & 53.80 & 48.07 & 54.91 & 43.42 & 63.04 \\
        \bottomrule
    \end{tabular}
    \caption{Zero-shot evaluation results for each source examination. NUAT HS1, HS2, and HS3 refer to high school 1st, 2nd, and 3rd grades, while G9 and G7 in CSE denote 9th and 7th grades, respectively.}
    \label{table_per_source}
\end{table*}

\begin{table}[t!]
\setlength{\tabcolsep}{0.85em} 
\renewcommand{\arraystretch}{1.0} 
\centering
\begin{tabular}{@{}c|c|c@{}}
\toprule
Exam      & years         & \begin{tabular}[c]{@{}c@{}}No. QA pairs\end{tabular} \\ \midrule
CSAT               & 1994-2023             & 70                       \\
NUAT               & 2006-2024             & 861                       \\
QE                 & 2001-2024            & 178                       \\
CSE               & 2006-2023           & 415                       \\ \midrule
\multicolumn{2}{c|}{Total} & 1,524        \\  
\toprule
\end{tabular}
\caption{The source data collection years.}  
\label{tab:source_data_details} %
\end{table}

\subsection{Data Categorization}
We classify the preprocessed data into one of 16 subcategories by three Korean language majors. 
Each annotator independently classifies each question, first identifying the main linguistic category and then assigning the appropriate subcategory. Labels are finalized by majority vote, and disagreements are resolved through discussion. 
We remove questions requiring knowledge from more than three categories to focus on the evaluation of each linguistic subcategory.
Considering the minimum hourly wage, we compensated data annotators approximately \$0.15 per question.

\begin{table*}[t!]
    \centering
    \begin{tabular}{l|ccccc}
        \toprule
        Model & \multicolumn{1}{l}{Class 1 (\%)} & \multicolumn{1}{l}{Class 2 (\%)} & \multicolumn{1}{l}{Class 3 (\%)} & \multicolumn{1}{l}{Class 4 (\%)} & \multicolumn{1}{l}{Class 5 (\%)} \\
        \midrule
        HyperClova-HCX-003 & 14.3 & 24.7 & 29.1 & 21.50 & 10.3 \\
        Claude-3.5-Sonnet & 19 & 25.5 & 23.9 & 21.7 & 9.9 \\
        GPT-4o & 16.2 & 24 & 25.1 & 22.00 & 12.7 \\
        \midrule
        Gold Label & 22.4 & 23.3 & 21.5 & 21.00 & 11.8\\
        \bottomrule
    \end{tabular}
    \caption{Class distribution of model predictions on the KoGEM benchmark. Each row shows the percentage of predictions falling into each class for the respective model. The gold label row represents the actual distribution of labels in the dataset, providing a reference for comparison.}
    \label{tab:class_distribution}
\end{table*}

\subsection{Description for Subcategories}
\label{sec:appendix_taxonomy}
Table~\ref{table:taxonomy_descriptions} presents a comprehensive breakdown of the linguistic categories, their subcategories, and corresponding descriptions. The framework is organized into five categories: \texttt{Phonology, Morphology, Syntax, Semantics}, and \texttt{Norms}. Each main category is further divided into subcategories.

First, the Phonology category encompasses the \texttt{Phonological System}, which refers to the system of Korean phonemes, and \texttt{Phonological Alternation}, which describes phenomena such as the insertion, deletion, and replacement of phonemes in specific environments. An example of phonological replacement can be found in the word `국물\textsubscript{soup}.' Specifically, while `국물\textsubscript{soup}' is composed of the syllables `국/guk/' and `물/mul/', its pronunciation is realized as [gung.mul], not /guk.mul/ due to nasalization. It is common for certain phonemes to change or be deleted depending on their phonological environment.

Second, the Morphology category covers aspects such as \texttt{Part-of-Speech}, \texttt{Morphemes}, and \texttt{Word Formation}. Unlike part-of-speech classification in English, which involves eight parts of speech, Korean has nine parts of speech (Table~\ref{table:taxonomy_descriptions}).
Additionally, the questions in the \texttt{Morpheme} subcategory require knowledge of the concepts and types of morphemes, such as lexical morphemes that carry core meanings (e.g., `꽃\textsubscript{flower}', `해\textsubscript{sun}') and functional morphemes that serve grammatical functions (e.g., `의\textsubscript{of}', `고\textsubscript{and}').
To solve the \texttt{Word Formation} task, both models and humans should understand Korean word formation rules, such as how compound words (e.g., `식기+세척기\textsubscript{dish+washer}') and derived words (e.g., `씻기\textsubscript{washing}' consists of `씻-\textsubscript{wash}' + `-기\textsubscript{-ing}') are formed. They should be able to identify the morphemes that make up each word.

Third, the Syntax category focuses on the syntactic structures of Korean sentences. We further include knowledge of syntactic features, which span grammatical components of sentences, such as tense (e.g., prefix ending `겠\textsubscript{future tense}') and negative adverbs (e.g., `안\textsubscript{not}' and `못\textsubscript{unable}'). The Semantics category examines the meaning of words, sentences, and the context of conversations. In the \texttt{Vocabulary} subcategory, we attempt to evaluate rote knowledge of words, which refers to their dictionary definitions. The questions in the \texttt{Lexical Semantics} subcategory assess the ability to identify word relationships, such as the connection between `orange' and `fruit'.

Lastly, the Norm category includes subcategories such as \texttt{Orthography}, \texttt{Standard Pronunciation}, \texttt{Loanword Orthography}, and \texttt{Romanization}, providing rules for accurate language use. \texttt{Cross-Category} addresses interdisciplinary topics that involve multiple subcategories within the Norms category. For instance, a question such as \textit{``Which one is correctly written according to Korean orthography and standard language rules?''} is included in \texttt{Cross-Category}.

We further provide specific example QAs for the 16 subcategories of KoGEM from Figure~\ref{fig:example_phonology} to Figure~\ref{fig:example_norms}. This serves as a detailed reference for understanding the linguistic elements assessed in KoGEM, highlighting its comprehensive and systematic design.

\begin{figure*}[t!]
    \centering
    \includegraphics[width=0.98\linewidth]{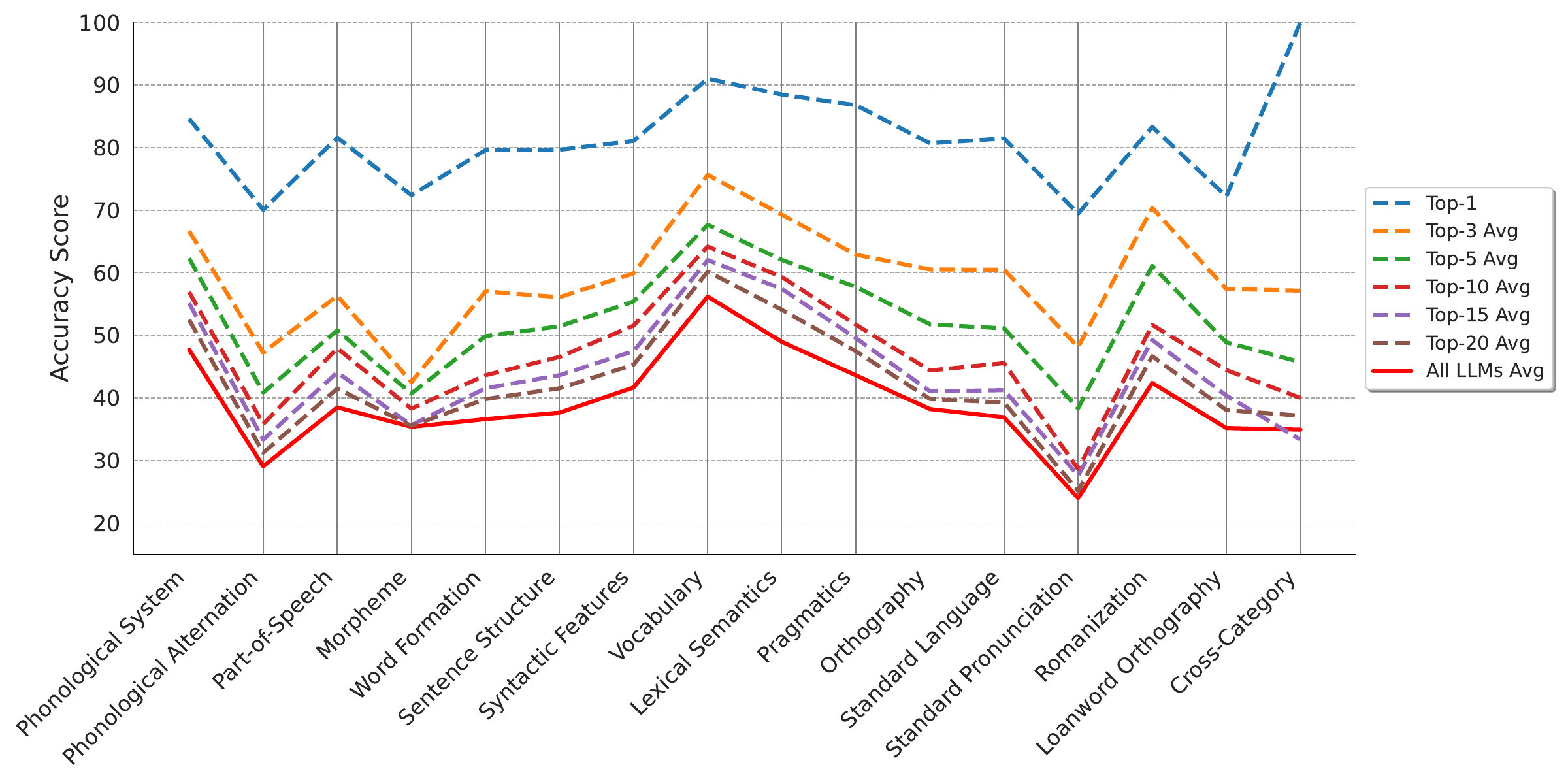}
    \caption{Accuracy scores across subcategories for different Top-k average settings and overall LLM performance.}
    \label{fig:subcategory_cumulative}
\end{figure*}

\section{Results per Data Source}
\label{sec:appendix_results_per_data_source}
Considering the varying levels of difficulty across the different exams from which our benchmark was derived, we conducted separate evaluations of LLMs' and humans' performance for each source dataset. The results of these evaluations are presented in Table~\ref{table_per_source}, and the category distribution by source is shown in Figure~\ref{fig:subcategory_numbers}.

\paragraph{Distribution} 
Among the nine source datasets, NUAT and CSAT contain a relatively high proportion of phonology, morphology, and syntax questions, with a minimal proportion of norms questions. In contrast, CSEs feature a notably larger share of norms and semantics questions. This variation in question type is also reflected in the results: in particular, humans scored higher on NUAT and CSAT than on CSEs. These findings suggest that humans outperform LLMs in phonology, morphology, and syntax tasks, which require empirical knowledge, reasoning, and intuition, compared to vocabulary and rigid norms that rely more on memorization.

\paragraph{Difficulty} 
As illustrated in Figure~\ref{fig:subcategory_numbers}, NUATs generally exhibit a consistent distribution of question types, but their difficulty increases as students progress through grades. This pattern is reflected in LLM evaluation results: models performed best on the first-year high school mock exam but showed a steady decline in performance with advancing grades, reaching their lowest scores on CSAT. Additionally, a comparable trend was observed in CSEs. While the types of questions remained broadly consistent, the 9th-grade exam is regarded as easier, whereas the 7th-grade exam is notably more challenging, leading to a corresponding decrease in LLM performance. Conversely, on HSQE, which requires comparatively less intensive study, most LLMs achieved their highest scores, further highlighting the correlation between exam difficulty and LLM performance.

\begin{table*}[t!]
    \setlength{\tabcolsep}{4.5pt}             
    \renewcommand{\arraystretch}{1.00}       
    \small
    \begin{tabular}{l|ll|ccccccc}
        \toprule
        Language & \multicolumn{2}{c|}{Model} & Phonology & Morphology & Syntax & Semantics & Norm & Avg. & 
        $\Delta$ \\ 
        \midrule
        \multirow{9}{*}{Korean} & \multirow{3}{*}{EEVE-Instruct-10.8B-v1.0} & 0-shot & 22.54 & 27.24 & 27.85 & 40.78 & 27.73 & 30.25 & - \\
         &  & 1-shot & 21.23 & 29.70 & 26.96 & 38.52 & 29.95 & 29.97 & -0.28 \\
         &  & 5-shot & 23.15 & 30.23 & 25.48 & 37.95 & 24.64 & 29.00 & -1.25 \\ \cmidrule{2-10} 
         & \multirow{3}{*}{EXAONE-3.5-7.8B-Instruct} & 0-shot & 24.88 & 30.22 & 32.19 & 43.64 & 31.36 & 33.60 & - \\
         &  & 1-shot & 28.30 & 31.20 & 32.41 & 43.06 & 33.64 & 34.49 & 0.89 \\
         &  & 5-shot & 27.72 & 30.87 & 33.84 & 41.11 & 30.82 & 33.86 & 0.26 \\ \cmidrule{2-10}
         & \multirow{3}{*}{EXAONE-3.5-32B-Instruct} & 0-shot & 27.23 & 37.31 & 36.30 & 50.65 & 37.27 & 38.98 & - \\
         &  & 1-shot & 29.58 & 37.07 & 37.03 & 50.97 & 40.87 & 39.97 & 0.99 \\
         &  & 5-shot & 28.70 & 37.13 & 38.44 & 52.44 & 41.13 & 40.52 & 1.54 \\ 
         \midrule
        \multirow{12}{*}{English} & \multirow{3}{*}{Qwen2.5-14B-Instruct} & 0-shot & 29.58 & 38.43 & 39.27 & 52.21 & 33.64 & 40.22 & - \\
         &  & 1-shot & 26.42 & 36.84 & 37.10 & 53.30 & 34.56 & 39.26 & -0.96 \\
         &  & 5-shot & 27.09 & 40.70 & 37.98 & 51.94 & 32.37 & 39.61 & -0.61 \\ \cmidrule{2-10}
         & \multirow{3}{*}{Qwen2.5-32B-Instruct} & 0-shot & 26.29 & 36.19 & 41.78 & 62.86 & 35.45 & 43.04 & - \\
         &  & 1-shot & 28.30 & 38.72 & 43.55 & 63.85 & 33.18 & 44.16 & 1.12 \\
         &  & 5-shot & 30.05 & 39.53 & 47.36 & 64.72 & 32.85 & 45.78 & 2.74 \\ \cmidrule{2-10}
         & \multirow{3}{*}{Claude-3.5-Sonnet} & 0-shot & 47.42 & 52.61 & 64.38 & 74.55 & 46.82 & 59.97 & - \\
         &  & 1-shot & 48.45 & 54.50 & 65.22 & 75.12 & 48.58 & 60.99 & 1.02 \\
         &  & 5-shot & 49.01 & 56.37 & 66.99 & 75.54 & 49.51 & 62.23 & 2.26 \\ \cmidrule{2-10}
         & \multirow{3}{*}{GPT-4o} & 0-shot & 44.60 & 51.49 & 55.48 & 71.95 & 58.64 & 57.87 & - \\
         &  & 1-shot & 45.02 & 53.56 & 59.30 & 73.56 & 58.26 & 59.75 & 1.88 \\
         &  & 5-shot & 51.24 & 50.20 & 61.35 & 74.03 & 59.91 & 60.94 & 3.07 \\ 
         \bottomrule
    \end{tabular}
    \caption{Few-shot accuracy evaluation results on our KoGEM benchmark. It consists of two segments: Korean-centric LLMs trained mainly on Korean data, and English-centric LLMs trained primarily on English data, respectively. $\Delta$ denotes the difference between 0-shot and n-shot accuracy.}
    \label{tab:few-shot-results}
\end{table*}

\section{Top-k Performance Variation Across Subcategories}
\label{appdx:cum_zero_shot_results}

Figure~\ref{fig:subcategory_cumulative} displays the zero-shot accuracy scores across a range of subcategories under various Top-k averaging conditions (Top-1 through Top-20), along with the overall average across all LLMs. While the absolute accuracy values vary with different Top-k settings, the relative performance trends among the linguistic subcategories remain largely consistent. This stability in ranking across Top-k variants justifies the use of the all LLMs average score as a representative metric in the main analysis in Figure~\ref{fig:subcategory_results}. It allows us to abstract away from Top-k variance while retaining the comparative insight across categories.

\section{Comparison of Predicted and Gold Class Distributions on KoGEM}

We analyze the distribution of predicted classes produced by each model on the KoGEM benchmark to examine potential discrepancies or biases in class prediction frequencies. Table~\ref{tab:class_distribution} summarizes the relative frequencies of predictions for five predefined classes by three LLMs, such as HyperClova-HCX-003, Claude-3.5-Sonnet, and GPT-4o, along with the ground truth class distribution.

The results reveal distinct distributional tendencies across models. For instance, HyperClova-HCX-003 exhibits a higher frequency for Class 3 (29.1\%) and a lower frequency for Class 1 (14.3\%) compared to the gold distribution. Claude-3.5-Sonnet and GPT-4o produce more balanced distributions, though certain classes, such as Class 5, show slight over- or under-predictions.

Despite these model-specific differences, the overall predicted distributions remain relatively close to the gold label distribution. This indicates that the models are generally able to approximate the class frequencies present in the dataset, suggesting no severe class imbalance or systematic deviation in their outputs.

\section{Few-shot Evaluation of KoGEM}
\label{appx:few-shot_results}
To explore the effect of exemplar-based context in few-shot learning, we conducted additional 1-shot and 5-shot evaluation experiments. We used three Korean-centric LLMs, such as EEVE-v1.0-10.8B-Instruct, EXAONE-3.5-7.8B-Instruct, and EXAONE-3.5-32B-Instruct, and four English-centric LLMs, such as Qwen2.5-14B-Instruct, Qwen2.5-32B-Instruct, Claude-3.5-Sonnet, and GPT-4o. As shown in Table~\ref{tab:few-shot-results}, the results showed that LLMs with at least 32B parameters exhibited notable improvements, indicating that exemplar-based context enhances performance on tasks requiring nuanced understanding. In contrast, smaller models struggled to effectively utilize the provided examples.

\begin{table*}[t!] 
    \renewcommand{\arraystretch}{1.1}    
    \centering
        \begin{tabular}{>{\raggedright\arraybackslash}m{3.5cm}|>{\raggedright}m{5.3cm}|>{\raggedright\arraybackslash}p{6cm}}
        \toprule
        \multicolumn{1}{c|}{Linguistic Category} & \multicolumn{1}{l|}{Subcategory} & \multicolumn{1}{c}{Description} \\
        \midrule
        \multicolumn{1}{c|}{\multirow{6}{*}{Phonology}} & \multirow{3}{*}{Phonological System (PHS)} & The system of Korean consonants and vowels and the principles of syllable formation. \\
        \cline{2-3}
        & \multirow{3}{*}{Phonological Alternation (PHA)} & Phonological insertion, deletion, replacement, and combination phenomena in Korean. \\
        \hline
        \multicolumn{1}{c|}{\multirow{9}{*}{Morphology}} & \multirow{4}{*}{Part-of-Speech (POS)} & The nine parts of speech in Korean (i.e., noun, pronoun, numeral, verb, adjective, determiner, adverb, particle, exclamation). \\
        \cline{2-3}
        & \multirow{3}{*}{Morpheme (MOR)} & The concept and types of morphemes (e.g., lexical morphemes, functional morphemes). \\
        \cline{2-3}
        & \multirow{2}{*}{Word Formation (WOF)} & Word formation process such as compounding and derivation. \\
        \hline
        \multicolumn{1}{c|}{\multirow{6}{*}{Syntax}} & \multirow{3}{*}{Sentence Structure (STS)} & The sentence structure and the components of a sentence (e.g., subject, predicate, object, SVO word order, etc.). \\
        \cline{2-3}
        & \multirow{3}{*}{Syntactic Features (STF)} & Various syntactic elements in sentences (e.g., tense, particles, negation, voice, honorifics, etc.). \\
        \hline
        \multicolumn{1}{c|}{\multirow{7}{*}{Semantic}} & \multirow{2}{*}{Vocabulary (VOC)} & Rote knowledge of the dictionary definition of words. \\
        \cline{2-3}
        & \multirow{3}{*}{Lexical Semantics (LES)} & Understanding semantic relationships of words (e.g., homonym, antonym, polysemy, etc.). \\
        \cline{2-3}
        & \multirow{2}{*}{Pragmatics (PRA)} & Understanding discourse (e.g., context, reference, intentions of speakers, etc.). \\
        \hline
        \multicolumn{1}{c|}{\multirow{12}{*}{Norms}} & \multirow{1}{*}{Orthography (ORT)} & Accurate rules of Korean spelling. \\
        \cline{2-3}
        & \multirow{2}{*}{Standard Language (STL)} & Accurate knowledge of standard Korean. \\
        \cline{2-3}
        & \multirow{2}{*}{Standard Pronunciation (STP)} & Accurate rules of Korean pronunciation. \\
        \cline{2-3}
        & \multirow{2}{*}{Loanword Orthography (LWO)} & Korean orthography rules for loanwords. \\
        \cline{2-3}
        & \multirow{2}{*}{Romanization (ROM)} & Rules for transcribing Korean into Roman script. \\
        \cline{2-3}
        & \multirow{3}{*}{Cross-Category (CRC)} & Questions asking about two or more subcategories within the Norms category. \\
        \bottomrule
    \end{tabular}
    \caption{Description of each subcategory. The three characters (e.g., PHS, PHA) in parentheses next to each subcategory represent its abbreviation.} 
    \label{table:taxonomy_descriptions} 
\end{table*}

\begin{table*}[t!]
\centering
\tiny
\setlength{\tabcolsep}{2pt}               
\renewcommand{\arraystretch}{1.4}           
\begin{tabular}{ccl|cc|ccc|cc|ccc|cccccc|c}
\toprule
\multicolumn{1}{c|}{\multirow{2}{*}{Language}} & \multicolumn{1}{c|}{\multirow{2}{*}{Type}} & \multicolumn{1}{c|}{\multirow{2}{*}{Model}} & \multicolumn{2}{c|}{Phonology} & \multicolumn{3}{c|}{Morphology} & \multicolumn{2}{c|}{Syntax} & \multicolumn{3}{c|}{Semantics} & \multicolumn{6}{c|}{Norms} & \multirow{2}{*}{Total} \\ \cline{4-19}
\multicolumn{1}{c|}{} & \multicolumn{1}{c|}{} & \multicolumn{1}{c|}{} & PHS & PHA & POS & MOR & WOF & STS & STF & VOC & LES & \multicolumn{1}{c|}{PRA} & ORT & STL & STP & LWO & ROM & \multicolumn{1}{c|}{CRC} &  \\ 
\midrule
\multicolumn{1}{c|}{\multirow{8}{*}{Korean}} & \multicolumn{1}{c|}{\multirow{6}{*}{Open}} & llama-3-Korean-Bllossom-8B & 42.31 & 21.93 & 25.29 & 31.03 & 23.68 & 19.31 & 27.70 & 37.72 & 30.91 & 35.85 & 33.33 & 29.63 & 22.22 & 27.78 & 22.22 & 14.29 & 27.10 \\
\multicolumn{1}{c|}{} & \multicolumn{1}{c|}{} & SOLAR-10.7B-Instruct-v1.0 & 30.77 & 24.06 & 24.14 & 31.03 & 25.66 & 24.83 & 30.41 & 38.32 & 28.48 & 20.75 & 31.58 & 14.81 & 16.67 & 16.67 & 33.33 & 14.29 & 27.36 \\
\multicolumn{1}{c|}{} & \multicolumn{1}{c|}{} & KULLM3 & 15.38 & 22.46 & 31.03 & 31.03 & 23.03 & 25.86 & 26.35 & 34.13 & 26.67 & 22.64 & 32.46 & 29.63 & 13.89 & 33.33 & 22.22 & 28.57 & 26.64 \\
\multicolumn{1}{c|}{} & \multicolumn{1}{c|}{} & EEVE-Korean-Instruct-10.8B-v1.0 & 26.92 & 21.93 & 26.44 & 37.93 & 25.66 & 26.90 & 29.73 & 50.30 & 32.12 & 37.74 & 29.82 & 25.93 & 19.44 & 22.22 & 33.33 & 42.86 & 30.25 \\
\multicolumn{1}{c|}{} & \multicolumn{1}{c|}{} & EXAONE-3.5-7.8B-Instruct & 38.46 & 22.99 & 34.48 & 31.03 & 27.63 & 28.97 & 38.51 & 50.30 & 38.79 & 37.74 & 35.09 & 29.63 & 22.22 & 38.89 & 22.22 & 28.57 & 33.60 \\
\multicolumn{1}{c|}{} & \multicolumn{1}{c|}{} & EXAONE-3.5-32B-Instruct & 38.46 & 25.67 & 35.63 & {\ul 51.72} & 35.53 & 32.76 & 43.24 & 58.08 & 46.06 & 41.51 & 37.72 & 44.44 & 25.00 & 50.00 & 33.33 & 42.86 & 38.98 \\ 
\cmidrule{2-20} 
\multicolumn{1}{c|}{} & \multicolumn{1}{c|}{\multirow{2}{*}{Closed}} & HyperClova-HCX-DASH-001 & 46.15 & 20.86 & 32.18 & 31.03 & 30.92 & 26.90 & 22.97 & 44.91 & 36.97 & 32.08 & 35.09 & 37.04 & 25.00 & 22.22 & 22.22 & 28.57 & 30.77 \\
\multicolumn{1}{c|}{} & \multicolumn{1}{c|}{} & HyperClova-HCX-003 & 46.15 & 30.48 & 48.28 & 31.03 & 40.13 & 38.97 & 45.27 & 64.07 & 53.94 & 32.08 & 52.63 & 66.67 & 25.00 & 50.00 & 33.33 & 57.14 & 44.62 \\ \midrule
\multicolumn{1}{c|}{\multirow{19}{*}{English}} & \multicolumn{1}{c|}{\multirow{10}{*}{Open}} & Gemma-2-9B-Instruct & 46.15 & 21.39 & 22.99 & 34.48 & 35.53 & 30.69 & 31.76 & 52.10 & 35.15 & 32.08 & 29.82 & 25.93 & 19.44 & 44.44 & 22.22 & 57.14 & 32.68 \\
\multicolumn{1}{c|}{} & \multicolumn{1}{c|}{} & Gemma-2-27B-Instruct & 34.62 & 26.20 & 31.03 & 24.14 & 29.61 & 37.24 & 35.81 & 50.30 & 43.03 & 50.94 & 29.82 & 25.93 & 13.89 & 38.89 & 44.44 & 42.86 & 35.70 \\
\multicolumn{1}{c|}{} & \multicolumn{1}{c|}{} & Qwen2.5-7B-Instruct & 42.31 & 20.32 & 29.89 & 37.93 & 30.26 & 31.03 & 39.86 & 52.10 & 38.18 & 39.62 & 30.70 & 25.93 & 11.11 & 22.22 & 16.67 & 28.57 & 33.27 \\
\multicolumn{1}{c|}{} & \multicolumn{1}{c|}{} & Qwen2.5-14B-Instruct & 53.85 & 26.20 & 36.78 & 48.28 & 37.50 & 38.28 & 41.22 & 54.49 & 50.30 & 50.94 & 37.72 & 22.22 & 19.44 & 38.89 & 38.89 & 57.14 & 40.22 \\
\multicolumn{1}{c|}{} & \multicolumn{1}{c|}{} & Qwen2.5-32B-Instruct & 42.31 & 24.06 & 44.83 & 24.14 & 33.55 & 42.41 & 40.54 & 64.67 & 64.85 & 50.94 & 38.60 & 37.04 & 16.67 & 55.56 & 27.78 & 42.86 & 43.04 \\
\multicolumn{1}{c|}{} & \multicolumn{1}{c|}{} & DeepSeek-R1-Distill-Qwen-14B & 53.85 & 25.67 & 31.03 & 20.69 & 34.87 & 39.31 & 39.19 & 47.90 & 50.91 & 37.74 & 32.46 & 44.44 & 25.00 & 50.00 & 16.67 & 14.29 & 37.73 \\
\multicolumn{1}{c|}{} & \multicolumn{1}{c|}{} & DeepSeek-R1-Distill-Qwen-32B & 65.38 & 32.09 & 48.28 & 31.03 & 46.71 & 39.31 & 42.57 & 67.07 & 61.21 & 45.28 & 28.95 & 25.93 & 27.78 & 50.00 & 38.89 & 0.00 & 44.55 \\
\multicolumn{1}{c|}{} & \multicolumn{1}{c|}{} & \multicolumn{1}{l|}{s1-32B} & 53.85 & 37.97 & 40.23 & 44.83 & 44.74 & 44.14 & 50.00 & 65.87 & 61.82 & 54.72 & 44.74 & 33.33 & 33.33 & 33.33 & 50.00 & 14.29 & 48.03 \\
\multicolumn{1}{c|}{} & \multicolumn{1}{c|}{} & Llama-3.1-8B-Instruct & 42.31 & 21.93 & 17.24 & 31.03 & 26.97 & 23.79 & 24.32 & 43.11 & 29.09 & 37.74 & 27.19 & 22.22 & 25.00 & 33.33 & 27.78 & 28.57 & 27.62 \\
\multicolumn{1}{c|}{} & \multicolumn{1}{c|}{} & Llama-3-70B & 38.46 & 22.46 & 35.63 & 24.14 & 32.24 & 31.03 & 41.22 & 50.30 & 42.42 & 41.51 & 30.70 & 29.63 & 19.44 & 22.22 & 27.78 & 28.57 & 34.58 \\
\multicolumn{1}{c|}{} & \multicolumn{1}{c|}{} & Llama-3.1-405B & 57.69 & 34.76 & 49.43 & 44.83 & 39.47 & 47.24 & 46.62 & 65.27 & 63.64 & 49.06 & 32.46 & 44.44 & 19.44 & 44.44 & 55.56 & 42.86 & 47.18 \\ 
\cmidrule{2-20} 
\multicolumn{1}{c|}{} & \multicolumn{1}{c|}{\multirow{8}{*}{Closed}} & Gemini-1.5-flash & 50.00 & 35.83 & 47.13 & 41.38 & 33.55 & 41.38 & 50.68 & 69.46 & 56.97 & 45.28 & 43.86 & 37.04 & 11.11 & 44.44 & 33.33 & 0.00 & 45.34 \\
\multicolumn{1}{c|}{} & \multicolumn{1}{c|}{} & Gemini-2.0-flash-exp & 65.38 & {\ul 43.32} & 56.32 & 37.93 & 48.03 & 54.83 & 60.81 & 70.06 & {\ul 76.36} & 56.60 & 47.37 & 51.85 & 27.78 & 50.00 & 50.00 & {\ul 71.43} & 56.04 \\
\multicolumn{1}{c|}{} & \multicolumn{1}{c|}{} & Claude-3-haiku & 46.15 & 17.65 & 36.78 & 31.03 & 33.55 & 33.79 & 39.19 & 48.50 & 43.64 & 35.85 & 30.70 & 37.04 & 16.67 & 50.00 & 33.33 & 28.57 & 34.97 \\
\multicolumn{1}{c|}{} & \multicolumn{1}{c|}{} & Claude-3.5-Sonnet & {\ul 80.77} & 42.78 & {\ul 59.77} & 48.28 & 49.34 & {\ul 63.10} & {\ul 66.89} & 76.65 & 75.15 & {\ul 66.04} & 54.39 & 44.44 & 25.00 & {\ul 66.67} & 27.78 & 42.86 & {\ul 59.97} \\
\multicolumn{1}{c|}{} & \multicolumn{1}{c|}{} & GPT-3.5-turbo & 30.77 & 19.79 & 25.29 & 27.59 & 28.95 & 25.86 & 30.41 & 34.73 & 27.27 & 33.96 & 22.81 & 29.63 & 22.22 & 27.78 & 44.44 & 14.29 & 27.30 \\
\multicolumn{1}{c|}{} & \multicolumn{1}{c|}{} & GPT-4o-mini & 50.00 & 29.95 & 29.89 & 20.69 & 40.13 & 35.17 & 39.19 & 57.49 & 48.48 & 47.17 & 42.98 & 33.33 & 25.00 & 61.11 & 38.89 & 14.29 & 39.96 \\
\multicolumn{1}{c|}{} & \multicolumn{1}{c|}{} & GPT-4o & 65.38 & 41.71 & 57.47 & 34.48 & {\ul 51.32} & 53.45 & 59.46 & {\ul 78.44} & 70.91 & 54.72 & {\ul 57.89} & {\ul 66.67} & {\ul 50.00} & {\ul 66.67} & {\ul 61.11} & 57.14 & 57.87 \\
\multicolumn{1}{c|}{} & \multicolumn{1}{c|}{} & o1-preview & \textbf{84.62} & \textbf{70.05} & \textbf{81.61} & \textbf{72.41} & \textbf{79.61} & \textbf{79.66} & \textbf{81.08} & \textbf{91.02} & \textbf{88.48} & \textbf{86.79} & \textbf{80.70} & \textbf{81.48} & \textbf{69.44} & \textbf{83.33} & \textbf{72.22} & \textbf{100.00} & \textbf{81.04} \\ 
\midrule
\multicolumn{3}{c|}{LLMs Avg.} & 47.72 & 29.06 & 38.48 & 35.38 & 36.60 & 37.64 & 41.67 & 56.20 & 48.96 & 43.61 & 38.21 & 36.90 & 23.97 & 42.39 & 35.18 & 34.92 & 40.24 \\ 
\midrule
\multicolumn{3}{c|}{Human} & 65.13 & 66.93 & 50.67 & 62.28 & 59.19 & 64.12 & 65.88 & 70.86 & 69.35 & 76.46 & 55.30 & 50.25 & 58.46 & 60.20 & 41.68 & 51.58 & 63.04 \\ 
\bottomrule
\end{tabular}
\caption{Zero-shot evaluation results for the whole main linguistic category and subcategory.}
\label{tab:whole_results}
\end{table*}

\newpage

\begin{figure*}
    \centering
    \includegraphics[width=1.0\linewidth]{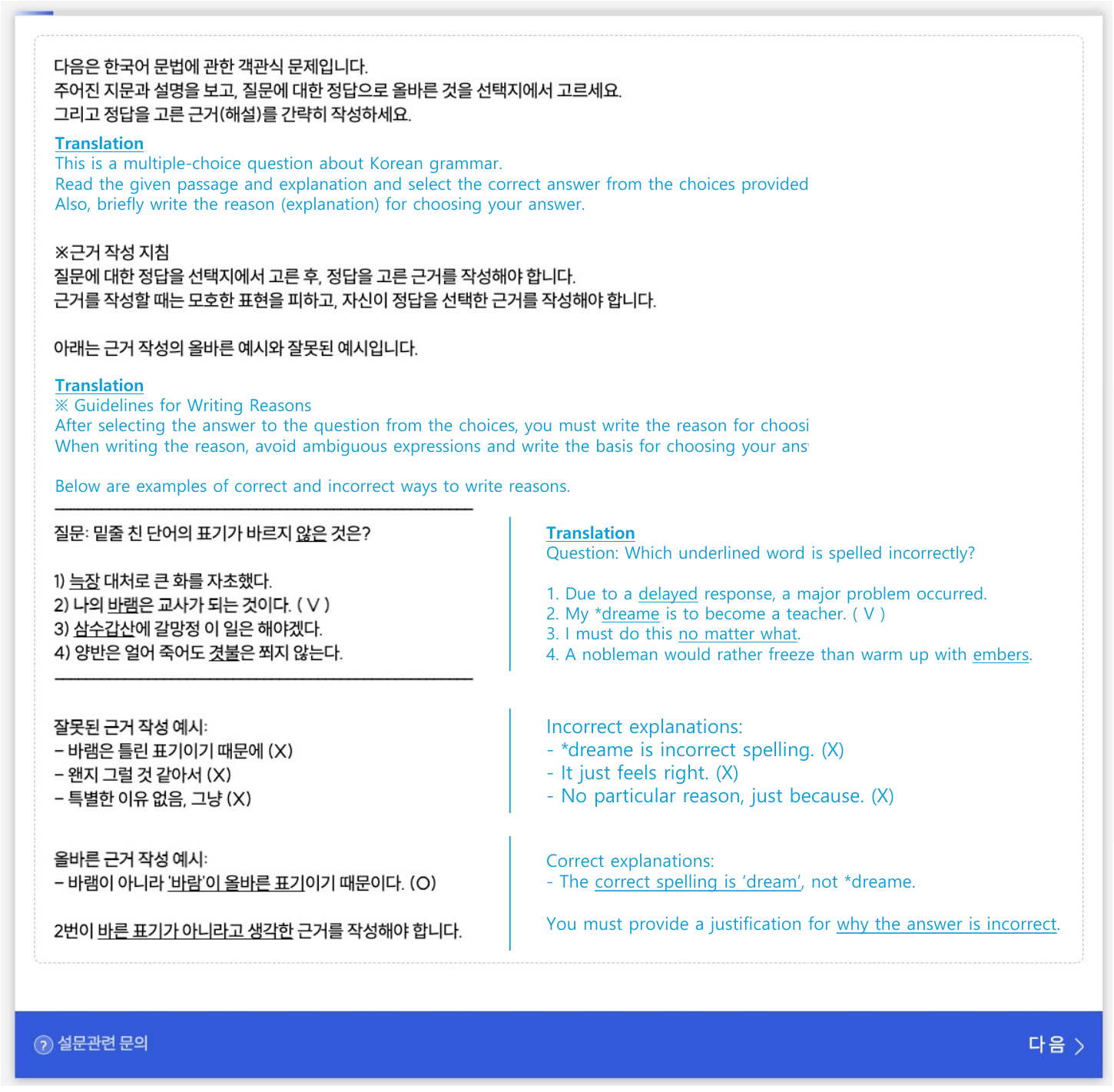}
    \caption{The instruction provided to crowdworkers for human evaluation outlines the evaluation process, task requirements, and criteria for rejecting invalid responses as incorrect. We have added English translation scripts in \textcolor{skyblue}{blue font} to clarify the meaning. However, these translation scripts were not provided to the crowdworkers.}
    \label{fig:human_instruction}
\end{figure*}

\begin{figure*}
    \centering
    \includegraphics[width=1.0\linewidth]{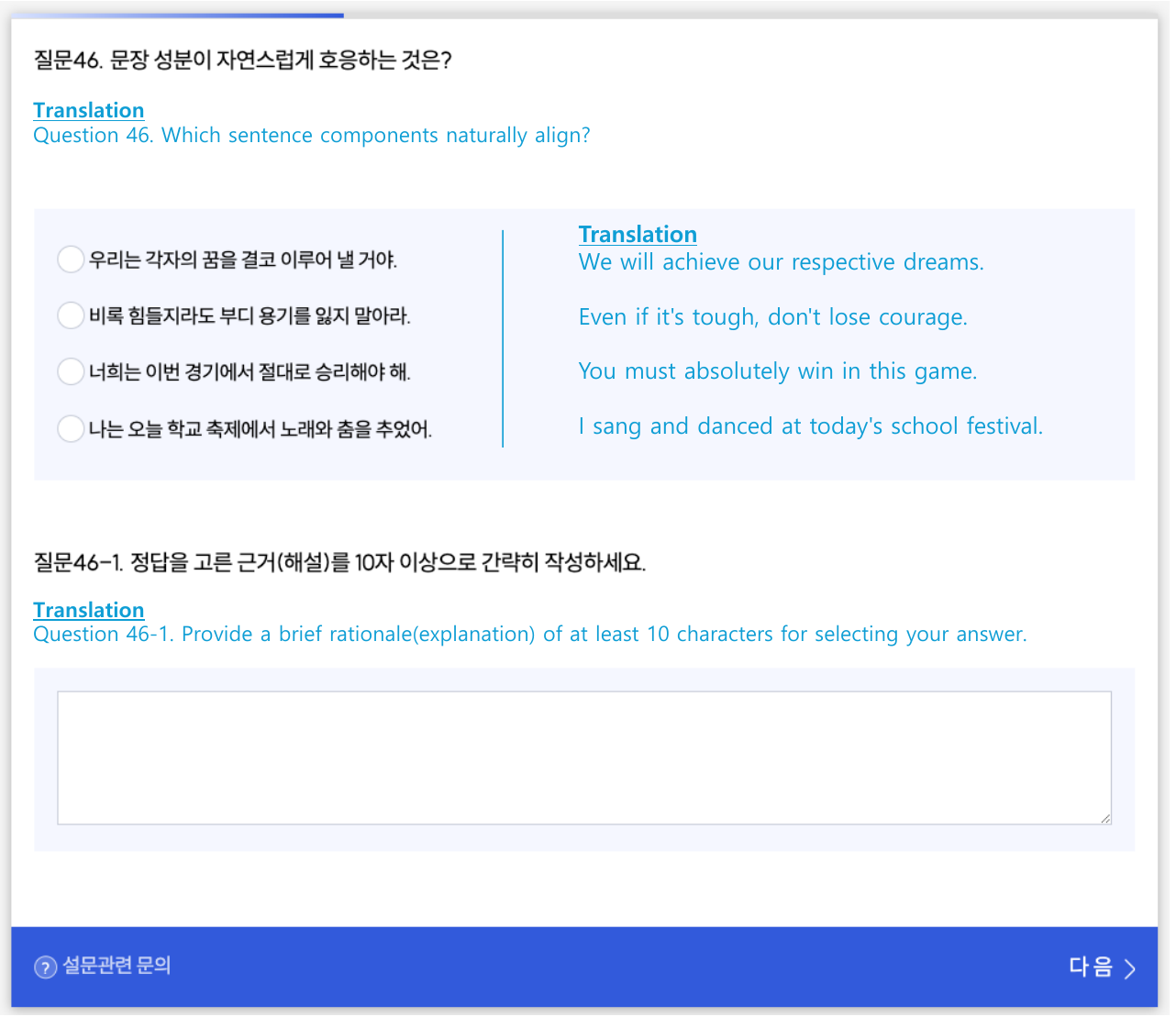}
    \caption{An example presented to participants during the human evaluation process illustrates the format and content of the exams. Specifically, it includes only \texttt{question} and \texttt{choices}, corresponding to T4 in Section~\ref{prompt_designs}. We have added English translation scripts in \textcolor{skyblue}{blue font} to clarify the meaning. However, these translation scripts were not provided to the crowdworkers.}
    \label{fig:human_test_sample}
\end{figure*}


\begin{figure*}[t!]
    \includegraphics[width=1.0\linewidth]{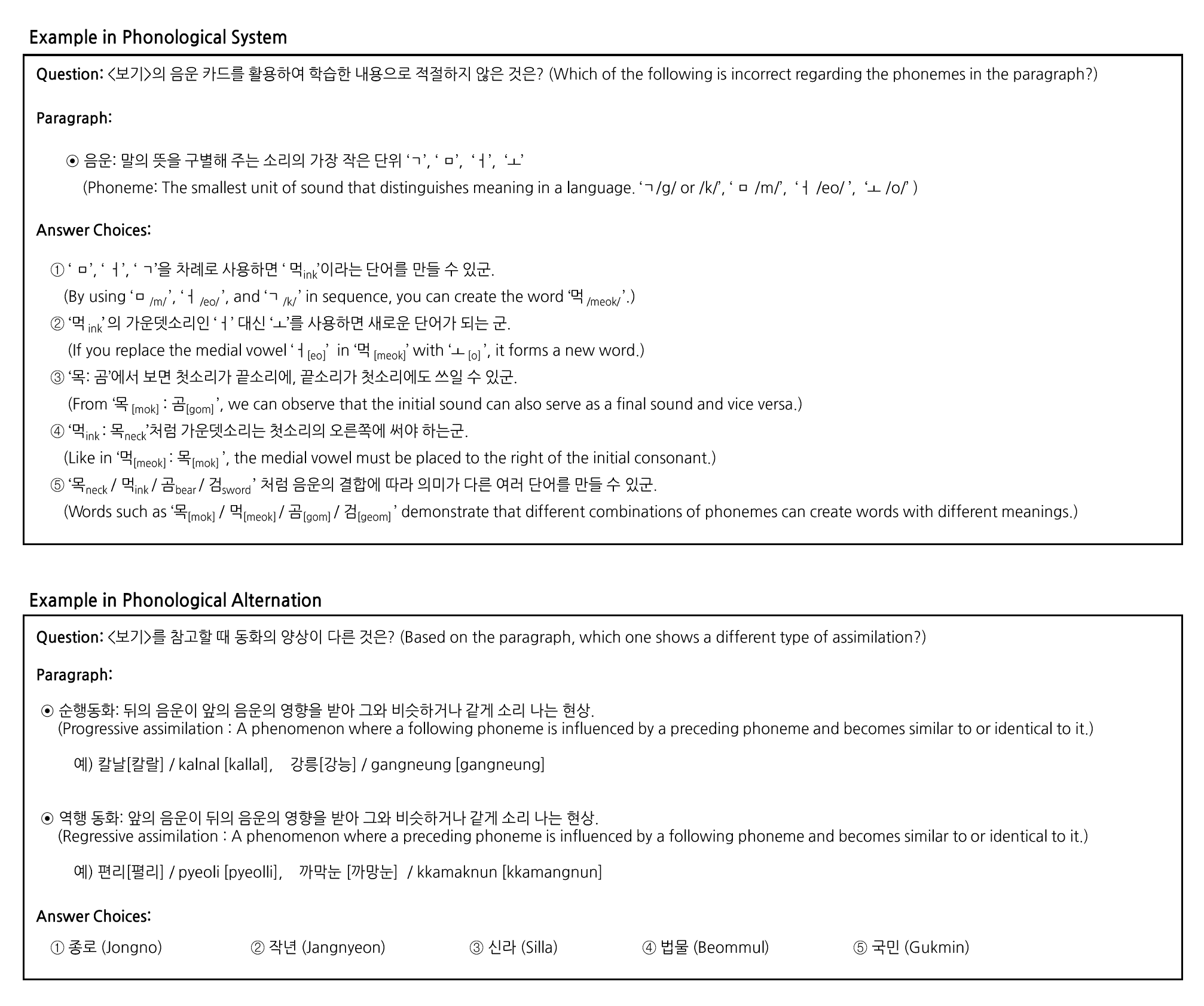}
    \caption{Examples in the \textit{Phonology} Category}
    \label{fig:example_phonology}
\end{figure*}

\begin{figure*}[t!]
    \includegraphics[width=1.0\linewidth]{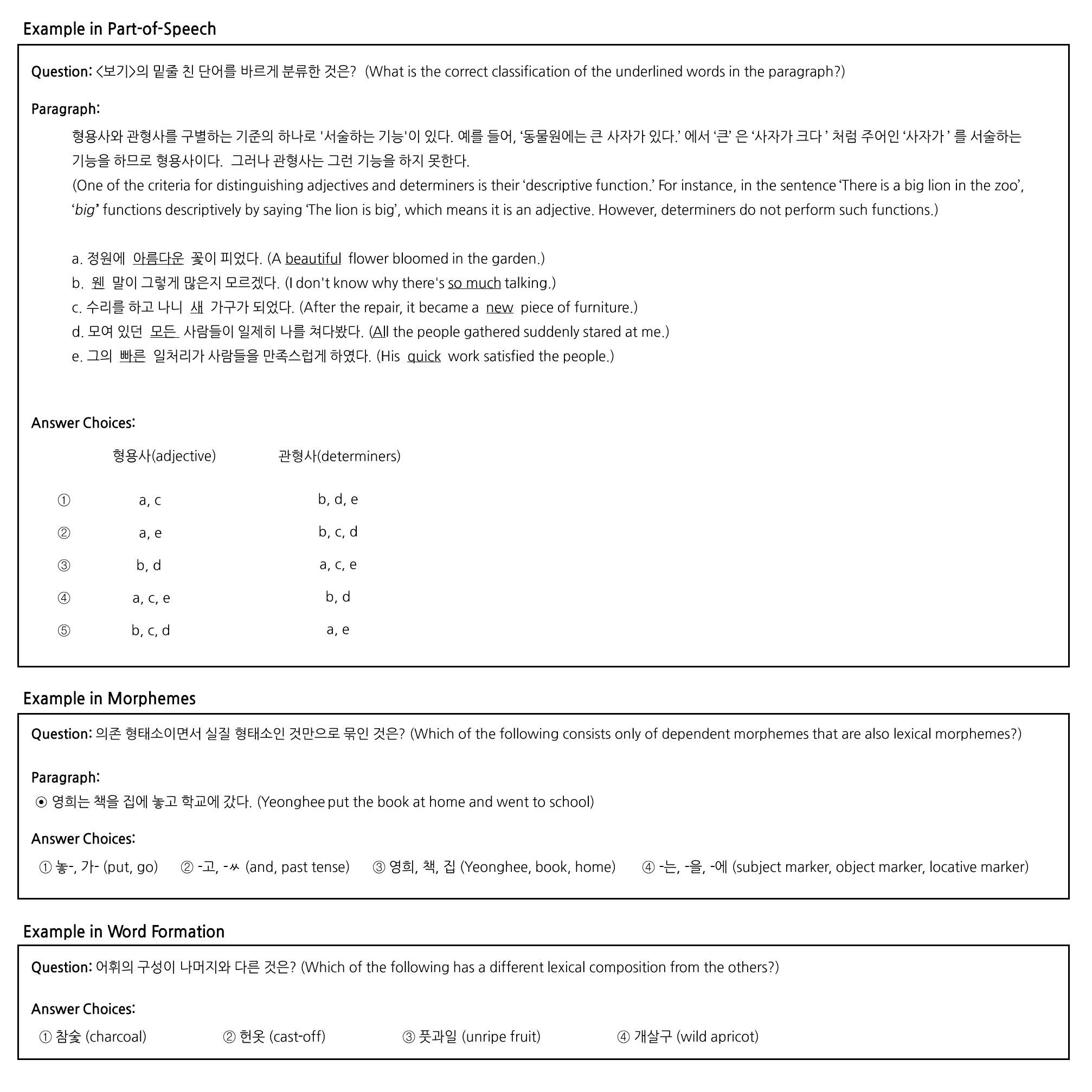}
    \caption{Examples in the \textit{Morphology} Category}
    \label{fig:example_morphology}
\end{figure*}

\begin{figure*}[t!]
    \includegraphics[width=1.0\linewidth]{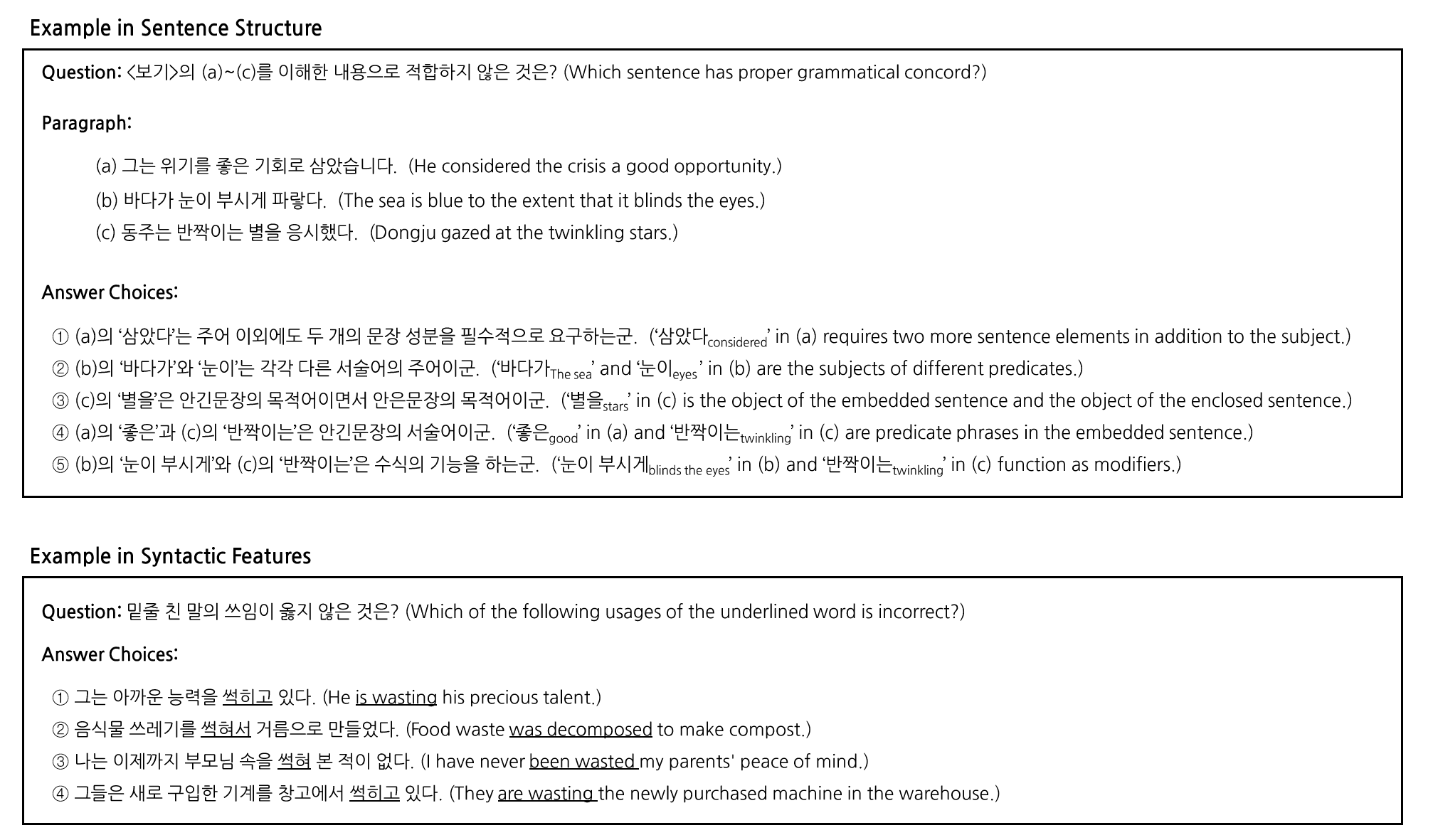}
    \caption{Examples in the \textit{Syntax} Category}
    \label{fig:example_syntax}
\end{figure*}

\begin{figure*}[t!]
    \includegraphics[width=1.0\linewidth]{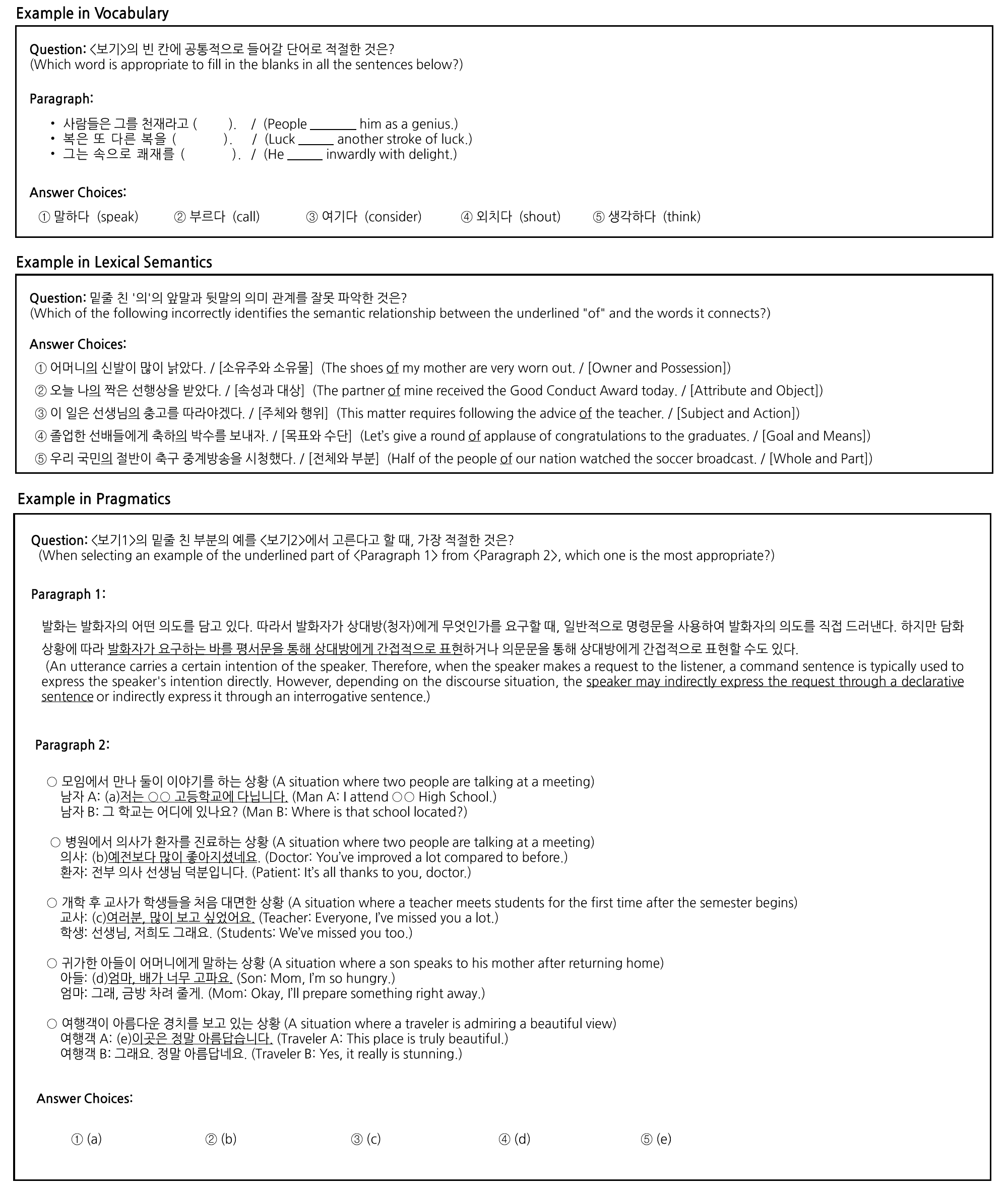}
    \caption{Examples in the \textit{Semantics} Category}
    \label{fig:example_semantics}
\end{figure*}

\begin{figure*}[t!]
    \includegraphics[width=1.0\linewidth]{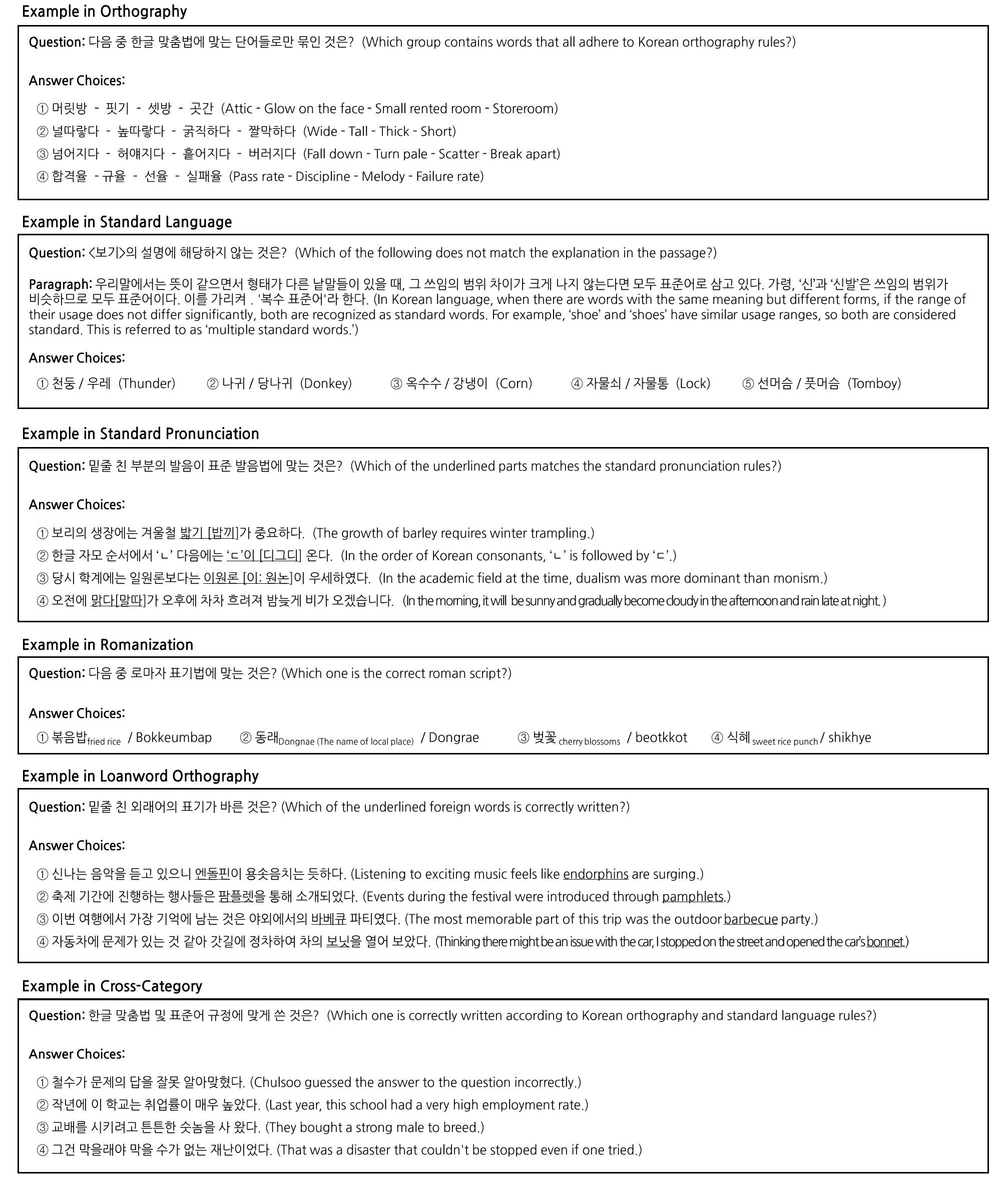}
    \caption{Examples in the \textit{Norms} Category}
    \label{fig:example_norms}
\end{figure*}

\end{CJK}

\end{document}